\documentclass[10pt]{article}
\newcommand{\documentdate}{29 V 2026}
\usepackage{a4wide,latexsym,graphicx,amsmath,amssymb,varioref}
\usepackage{graphics,xcolor,dsfont}

\DeclareMathAlphabet{\pazocal}{OMS}{zplm}{m}{n}
\newcommand{\calA}{{\pazocal{A}}}

\newcommand{\calF}{{\pazocal{F}}} 
\newcommand{\calI}{{\pazocal{I}}} 
\newcommand{\calJ}{{\pazocal{J}}} 
 
\newcommand{\calL}{{\pazocal{L}}} 
\newcommand{\calM}{{\pazocal{M}}} 
\newcommand{\calN}{{\pazocal{N}}} 
\newcommand{\calO}{{\pazocal{O}}}

\newcommand{\calS}{{\pazocal{S}}} 
\newcommand{\calU}{{\pazocal{U}}}

\newcommand{\beqn}[1]{\begin{equation}\label{#1}}
\newcommand{\eeqn}{\end{equation}}
\newcommand{\req}[1]{(\ref{#1})}
\newcommand{\ms}{\;\;\;\;}
\newcommand{\tim}[1]{\;\; \mbox{#1} \;\;}

\newtheorem{theorem}{Theorem}[section]
\newtheorem{assumption}{Assumption}
\newtheorem{lemma}[theorem]{Lemma}

\newcommand{\numsection}[1]{\section{#1}\setcounter{equation}{0}}
\newtheorem{corollary}[theorem]{Corollary}
\newcommand{\appnumsection}[1]{\section*{#1}
  \renewcommand{\theequation}{A.\arabic{equation}}
  \renewcommand{\thetheorem}{A.\arabic{theorem}}
  \renewcommand{\thetable}{A.\arabic{table}}
  \renewcommand{\thefigure}{A.\arabic{figure}}
  \renewcommand{\thesection}{A} }
\renewcommand{\theequation}{\arabic{section}.\arabic{equation}}
\renewcommand{\thefootnote}{(\arabic{footnote})}
\newcounter{algo}[section]
\renewcommand{\thealgo}{\thesection.\arabic{algo}}
\newcommand{\llem}[2]{\vspace{\baselineskip} 
\noindent\framebox[\textwidth]{\parbox{0.95\textwidth}{
\begin{lemma} \label{#1} \rm #2 \end{lemma} } } \vspace{\baselineskip} }
\newcommand{\llcor}[2]{\vspace{\baselineskip} 
\noindent\framebox[\textwidth]{\parbox{0.95\textwidth}{
\begin{corollary} \label{#1} \rm #2 \end{corollary} } } \vspace{\baselineskip} }
\newcommand{\algo}[3]{\refstepcounter{algo}
\begin{center}\begin{figure}[htbp]
\framebox[\textwidth]{
\parbox{0.95\textwidth} {\vspace{\topsep}
{\bf Algorithm \thealgo : #2}\label{#1}\\
\vspace*{-\topsep} \mbox{ }\\
{#3} \vspace{\topsep} }}
\end{figure}\end{center}}
\newcommand{\bpr}{{\bf Proof.} \hspace{1.5mm}}
\newcommand{\epr}{\hfill $\Box$ \vspace*{1em}}
\newcommand{\proof}[1]{
\begin{list}{}{
\setlength{\topsep}{0.0pt}
\setlength{\partopsep}{0.0pt}
\setlength{\leftmargin}{0.025\textwidth}
\setlength{\rightmargin}{0.5\leftmargin}
\setlength{\labelwidth}{0.5\leftmargin}
\setlength{\labelsep}{0.25\leftmargin}}
\item \bpr #1 \epr \noindent
\end{list}}
\newcommand{\lthm}[2]{\vspace{\baselineskip} 
\noindent\framebox[\textwidth]{\parbox{0.95\textwidth}{
\begin{theorem} \label{#1} \rm #2 \end{theorem} } } \vspace{\baselineskip} }
\newcommand{\ii}[1]{\{ 1, \ldots, #1 \}}
\newcommand{\iiz}[1]{\{ 0, \ldots, #1 \}}

\renewcommand{\Re}{\hbox{I\hskip -2pt R}}
\newcommand{\smallRe}{\hbox{\footnotesize I\hskip -2pt R}}
\newcommand{\bigfrac}[2]{\frac{\displaystyle #1}{\displaystyle #2}}
\newcommand{\bigsum}{\displaystyle \sum}

\newcommand{\sfrac}[2]{{\scriptstyle \frac{#1}{#2}}}
\newcommand{\kap}[1]{\kappa_{\mbox{\tiny #1}}}
\newcommand{\eqdef}{\stackrel{\rm def}{=}}
\newcommand{\half}{\sfrac{1}{2}}
\newcommand{\quarter}{\sfrac{1}{4}}
\newcommand{\flow}{f_{\rm low}}

\newcommand{\tG}{\widetilde{G}}

\newcommand{\ip}[1]{\left\langle#1\right\rangle}
\newcommand{\E}[1]{\mathbb{E}\!\left[#1 \right]}
\newcommand{\Econd}[2]{\mathbb{E}_{#1}\!\left[#2 \right]}

\newcommand{\hX}{\widehat{X}}
\newcommand{\hM}{\widehat{M}}

\newcommand{\hPi}{\widehat{\Pi}}
\newcommand{\hG}{\widehat{G}}
\newcommand{\mystack}[2]{_{\stackrel{\scriptstyle #1}{\scriptstyle #2}}}

\usepackage{booktabs}
\usepackage{multirow}

\DeclareMathOperator{\tr}{tr}

\definecolor{checkgreen}{rgb}{0,0.6,0}

\newif\ifcolorcomments
\colorcommentstrue

\newcommand{\comment}[1]{}

\newcommand{\al}[1]{{\small{\sf #1}}}
\newcommand{\tal}[1]{{\normalsize {\sf #1}}}
\newcommand{\lal}[1]{{\large {\sf #1}}}

\newcommand{\adanorm}{\tal{AdaNorm}}
\newcommand{\adagrad}{\tal{AdaGrad}}
\newcommand{\shampoo}{\tal{Shampoo}}
\newcommand{\muon}{\tal{Muon}}

\newcommand{\aname}{{\small {\sf ASADPREC}}}
\newcommand{\saname}{{\footnotesize {\sf ASADPREC}}}
\newcommand{\laname}{{\large {\sf ASADPREC}}}
\newcommand{\anameM}{{\small {\sf ASADPREC.M}}}
\newcommand{\anameMM}{{\small {\sf ASADPREC.MM}}}
\newcommand{\anameMG}{{\small {\sf ASADPREC.MG}}}
\newcommand{\asname}{{\small {\sf DADPREC}}}
\newcommand{\lasname}{{\large {\sf DADPREC}}}
\newcommand{\asnameMM}{{\small {\sf DADPREC.MM}}}
\newcommand{\asnameMG}{{\small {\sf DADPREC.MG}}}
\newcommand{\syncsingle}{{\small {\sf SyncSingle}}}

\newcommand{\syncmulti}{{\small {\sf SyncMulti}}}

\newcommand{\async}{\aname}
\newcommand{\sasync}{\saname}

\title{Stochastic convergence of parallel asynchronous adaptive first-order methods}

\author{
S. Gratton\thanks{Universit\'{e} de Toulouse, INP, IRIT, Toulouse, France. Email:
     serge.gratton@enseeiht.fr. Work partially supported by 3IA Artificial and
     Natural Intelligence Toulouse Institute (ANITI), French "Investing for the Future
     - PIA3" program under the Grant agreement ANR-19-PI3A-0004"}
~and Ph. L. Toint\thanks{Universit\'{e} de Toulouse, INP, IRIT,
  Toulouse, France, and NAXYS, University of Namur, Namur, Belgium. Email: philippe.toint@unamur.be}
}

\date{\documentdate}
\begin{document}

\renewcommand{\thefootnote}{\fnsymbol{footnote}}
\maketitle
\renewcommand{\thefootnote}{\arabic{footnote}}

\begin{abstract}
A new class of asynchronous adaptive first-order optimization methods
is introduced, comprising asynchronous variants of several popular
algorithms. Versions of these methods using momentum and/or inexact
normalization are also considered. The convergence of methods in the
class on non-convex functions is analyzed in a fully stochastic setting, and is shown to be
(up to logarithmic factors) of order $\calO(1/\sqrt{t})$ under
reasonable assumptions.  Numerical experiments suggest that such
asynchronous adaptive algorithms are very relevant in heterogeneous
large-scale machine learning systems.
\end{abstract}


\numsection{Introduction}
\cite{ZhuaWangLiuZhanLin19}
\cite{NablOyal22}

We consider the problem of finding a first-order critical point for
the optimization problem
\beqn{problem}
\min_{X\in\smallRe^N} F(X) = \E{f(X,\xi)}
\eeqn
where $f$ is a smooth, possibly nonconvex function of $X$ and $\xi$
is a suitably defined random variable. In other words, we are seeking
$X$ such that $\nabla_X \E{f(X,\xi)} = 0$.
Motivated by the growing importance of very large
models in deep learning, a substantial literature has
been devoted to the study of asynchronous algorithms for solving
\req{problem}. By far the most popular are stochastic
asynchronous steepest descent methods (also called asynchronous
gradient methods), whose modern efficient lock-free framework
has been pioneered by the \tal{Hogwild!} method \cite{NiuRechReWrig11}.
This method, like most of the methods we discuss in this paper,
maintains a vector $\hX$ of values for the problem's variables
in a memory shared by a number of processes/processors, each of which sequentially
\begin{enumerate}
\item selects a subset or block of the variables from $\hX$,
\vspace*{-2mm}
\item computes the gradient of $f$ at $\hX$ with respect to the variables of
  the block (this requires reading the vector $\hX$ from shared memory),
\vspace*{-2mm}
\item computes a gradient step (with method-specific stepsize) in the
  subspace spanned by this block using the block gradient, resulting
  in new values of the block variables,
\vspace*{-2mm}
\item updates the values of $\hX$ corresponding to the block with these
  new values.
\end{enumerate}
The whole operation is asynchronous because each process/processor does
not wait for other processes to complete their tasks (it is \emph{lock-free}). Of
course, this description is extremely synthetic and methods of the
type just described may still differ on a number of issues.
\begin{description}
\item[Geometry.] The first is to define how a particular block of variables is
  selected in Step~1. This selection can be random or deterministic,
  and is characterized by the size of the allowed ``overlap'' between
  blocks selected by different processes, ranging from totally
  separable (the blocks associated with running processes are
  disjoint), partially separable (limited overlap between running
  blocks is allowed --- for an introduction to partial separability,
  see \cite{GrieToin82c,GrieToin84a,ConnGoulToin92}), or 
  arbitrary (the size of the overlap is not constrained).
\vspace*{-2mm}
\item[Atomicity.] The next issue is the management of the read/write
  operations between local processes and shared memory.  Because of
  the asynchronous nature of the whole algorithm, it may happen
  that different processes attempt to read or write the same component
  of $\hX$ at the same time. To cope with this problem, it is generally
  assumed that the reading and writing operations are \emph{atomic} at
  some granularity level, meaning that reading/writing an atomic component of
  $\hX$ by a process cannot be interrupted by another process. We will
  see below that this concept also applies to other quantities than
  $\hX$, and that, in addition to purely hardware and message passing considerations,
  it also can be limited by the internal structure of the object to
  read/write, or by the norm used to measure it (for instance, objects like
  positive definite matrices need to maintain their positive definite nature).
\vspace*{-2mm}
\item[Consistency.]
  In addition, it may well happen that a process is reading parts of
  $\hX$ while other parts are being updated by other processes.  This
  blurs the notion of algorithm's iterates, because it is possible
  that a block gradient is computed for a value of the variables corresponding to an
  incomplete, ongoing update of $\hX$. Most algorithms (but not all) allow these
  \emph{inconsistent} read/writes.
\vspace*{-2mm}
\item[Delays.] In addition, the asynchrony of the whole algorithm
  implies that when a block gradient is computed, it uses the values
  of $\hX$ read from shared memory, but these values may result from
  writes (by other blocks) somewhat in the past.  The time difference
  between the moment $\hX$ is updated and the moment where it is used
  for computing a block gradient is called a \emph{delay}. The
  analysis of all   algorithms considered here imposes some kind of
  upper bound on such delays, either explicitly or in expectation.
\vspace*{-2mm}
\item[Choice of stepsize.] Finally, once the block gradient is computed,
  yielding a descent direction, a stepsize along this direction must
  be defined. Algorithms in the literature either assume that the
  stepsize is a ``suitably small'' constant, or that it is given by
  a (user) predefined decreasing sequence converging to zero.
\end{description}

Having brushed the landscape, we now review chronologically
significant proposals in the literature.
The pioneering \tal{Hogwild!} method proposed in
\cite{NiuRechReWrig11} was instrumental in introducing lock-free
variants of asynchronous steepest-descent. It applies to strongly
convex functions with partially separable geometry, for which block
gradients are computable.  Its stochastic convergence analysis assumes
that the gradient oracles are bounded and that the constant stepsize
can be chosen small enough compared to the gradient's Lipschitz
constant. Inconsistent reads are allowed.  This is not the case for
the \tal{AsySG-con} \cite{LianHuanLiLiu15} which, as the end of its
name suggests, insists on coherent reads, but applies to possibly
nonconvex functions with full-size gradients without any separability
requirement. Its analysis assumes bounded gradients and predefined
diminishing stepsizes. The \tal{ADrock} algorithm
\cite{PengXuYanYin16} is of a slightly different type, as it is better
viewed as a fixed point method, but it faces similar issues. It uses
full gradient vectors and requires that stepsizes decrease to
zero. Its analysis holds for convex functions with bounded gradients
but without any specific separability.  The \tal{KroMagnon}
algorithm \cite{Manietal17} is similar to \tal{Hogwild!} in that it is
a gradient-based method, allows block gradient evaluations and requires
a partially separable geometry. Establishing its rate of convergence
needs convex functions but no bound on the gradients is necessary. The
requirement on partially separable geometry is relaxed for the
versions of \tal{Hogwild!} studied in
\cite{NguyNguyvanDRichScheTaja18,NguyNguyRichScheTakavanD19} and
boundedness of the gradients is no longer requested. Instead of
assuming a small enough stepsize as in the original version, the
convergence analysis requires a stepsize decreasing to zero.
\tal{ASAGA} \cite{LeblPedrLaco18} is another gradient-based method
requiring, as is the case for the original \tal{Hogwild!}, partially
separable geometry, strongly convex functions and a sufficiently small
stepsize. However, there is no need to assume bounded gradients. The
method occasionally evaluates the full gradient (in shared memory) in
order to reduce the variance of the estimates. 
The \lal{FDG} algorithm of \cite{ZhuaWangLiuZhanLin19} partitions the
variables into blocks and interestingly
proposes to exploit the structure of backpropagation in neural
networks to compute the gradients with respect to the variables of a
block independently from those of another, in turns leading to an
asynchronous delayed-gradient method. The analysis stops with a
descent lemma requiring bounded gradients and knowledge of the
gradient's Lipschitz constant. 
The \tal{AsyFLEXA}
\cite{CannFaccKungScut20} deterministic algorithm is again slightly
different in that it relies on convex local models (instead of purely
linear ones as is the case for most gradient-based methods) and
applies to possibly nonconvex functions with block gradient
evaluations without separability requirement. Once more, the analysis
is carried out for small enough stepsize, but without assuming bounded
gradients.  The reference \cite{KungEganChatAlis21} discusses several
variants of a stochastic gradient-based method with momentum, as seen
in the context of continuous differential equations. We report here on
\al{PASSM}, a variant for potentially nonconvex totally separable
functions with block gradient evaluations, which appears to perform
best. Bounded gradients are required for its analysis and its
stepsizes follow a geometrically decreasing sequence. Interestingly, a
lock is required for writing the updated variables in shared memory.
One should also mention contributions of \cite{NadiMarkChatKung22} and
\cite{FeyzJoha23}. The first introduces the concept of ``elastic
consistency'', formalizing the conditions on delays for non-adaptive
gradient methods and showing that such conditions are necessary for
convergence of this type of methods. The second introduces general
recurrence occurring in many non-adaptive asynchronous stochastic
gradient methods and uses them to sharpen existing convergence
results, in particular by considering average delays rather than
maximum ones. Finally, the proposal of \cite{NablOyal22} considers an
asynchronous method for stricly convex functions with known Lipschitz
constant, putting special emphasis on communication costs. 

Table~\ref{tab:survey} \vpageref{tab:survey} provides an easy-to-read
summary of the above discussion, where column ``stoch.'' indicates whether
the method is stochastic,
and  ``cons. read'' indicates whether the method requires consistent
reading of shared memory. In the ``method type'' column, the
abbreviations ``convex approx'', ``prec. grad.'' and ``mom'' stand
for ``convex approximation'', ``preconditioned gradient'' and
``momentum'', respectively. Finally, the abbreviation ``adapt.'' in the
``stepsize'' column refers to adaptive stepsize, the technique we will
develop below.

\begin{table}
{\footnotesize
\hspace*{-10mm}\begin{tabular}{|c|c|c|c|c|c|c|c|c|c|}
\hline

&\multicolumn{2}{|c}{Problem}
&\multicolumn{4}{|c}{Algorithm's characteristics}
&\multicolumn{3}{|c|}{Assumptions}\\
\hline
&  convex  & separable & method & stoch. & shared  &  cons. & step &
bounded  & bounded  \\
&          & geometry  & type   &       &  info    &  read  & size &
gradients & delay   \\
\hline
\al{Hogwild!} \cite{NiuRechReWrig11}
& strong & partially & gradient & yes & $\hX$ & no &  small & yes &
yes \\
\hline
\al{AsySG-con} \cite{LianHuanLiLiu15}
& no &  no&  gradient & yes & $\hX$ & yes & $\searrow 0$  & no & yes \\
\hline
\al{``SGD''} \cite{ChatDuchRe15}
& yes & no & gradient & yes & $\hX$ & no &  $\searrow 0$  & yes & in
$\mathbb{E}$\\
\hline
\al{ARock} \cite{PengXuYanYin16}
& yes & no & fixed point  & yes & $\hX$ & no & small & yes & yes \\
\hline
\al{KroMagnon} \cite{Manietal17}
&  yes & partially & gradient & yes & $\hX$ & no & small & no & yes \\ 
\hline
\al{Hogwild!} \cite{NguyNguyvanDRichScheTaja18,NguyNguyRichScheTakavanD19}
 & strong & relaxed & gradient & yes & $\hX$ & no &  $\searrow 0$ &
no  & yes \\
\hline
\al{ASAGA} \cite{LeblPedrLaco18}
& strong & partially & gradient & yes & $\hX,(\hG)$ & no & small & no
& yes \\
\hline
\al{FDG} \cite{ZhuaWangLiuZhanLin19} & no & yes & gradient & yes & $\hX$ &
no & small & yes & yes\\
\hline
\al{AsyFLEXA} \cite{CannFaccKungScut20}
& no & no & convex approx. & no &  $\hX$ & no & small & no & yes \\
\hline
\al{PASSM} \cite{KungEganChatAlis21a,KungEganChatAlis21}
& no & yes & gradient+mom & yes & $\hX$ & no & $\searrow 0$ & in $\mathbb{E}$
& in $\mathbb{E}$\\
\hline
\al{DADAO} \cite{NablOyal22} &  yes & no & gradient & yes & $\hX$ & no
& small & no & yes \\
\hline
this paper
& no  & yes& prec. grad.+mom  &   yes & $\hX,(\hPi)$ & no & adapt.  & no
& yes \\
\hline
this paper
& no & yes & prec. grad.+mom &  yes & $\hX,(\hPi,\hM$) & no & adapt. & no &
yes\\
\hline
\end{tabular}
} 
\caption{\label{tab:survey}A summary of the relevant problem class, algorithm's characteristics and
conditions of analysis for stochastic asynchronous gradient
methods. Brackets around items in the ``shared info'' columns indicate
the items are not mandatory.
}
\end{table}

All the algorithms we have reviewed so far exhibit two (in our view,
significant) limitations. The first and most important one is that
pure gradient steps can lead to very slow convergence on (even
moderately) ill-conditioned problems. This issue has long been
addressed by the use of steepest-descent steps (a norm-dependent
concept, at variance with gradient steps) and (adaptive)
preconditioning (see \cite[Section~9.4]{BoydVand04}, for
instance). While these techniques have successfully been applied for
solving problem \req{problem} in the synchronous case, for instance in
methods like \tal{AdaGrad} \cite{DuchHazaSing11,McMaStre10,WardWuBott19}, \tal{Adam}
\cite{KingBa15}, \shampoo\ \cite{GuptKoreSing18,Vyasetal24} or
\muon\ \cite{Jordetal24,SiZhanShen25,ZhanLiuScha25} or in the distributed
case \cite{Liuetal25b}, they haven't yet (to the
authors' knowledge) been considered for asynchronous optimization.
The second limitation is that theoretical analysis of many of the published algorithms ignore the use
of momentum, despite its widespread nature (see \cite{Liuetal25b}, for
instance).

The main contribution of the present paper is to propose and analyze a
(to the authors' knowledge, first ever) class of algorithms containing
several \emph{stochastic asynchronous adaptively preconditioned
gradient methods}, possibly \emph{using momentum}, for
the nonconvex case. In particular, \adanorm, full and diagonal
\adagrad\ as well as adaptive variants of \shampoo\ and \muon\ are
covered.  Like for most recent proposals, bounded gradients are not
necessary for its analysis. The proposed algorithm builds on the
\tal{ADPREC} framework for (synchronous) stochastic first-order
methods \cite{GratToin26c}, which is itself based on the use of
general dual norms and covers a large class of stochastic adaptive
gradient-based methods. The present proposal takes advantage of its
powerful generality, and is defined in the geometric setting of
totally separable problems, as it is the case for \tal{PASSM}. While this
might be seen as theoretically restrictive, it still covers the very
important case of layer-wise preconditioning in training large neural
networks (see \cite{Yuetal17,Ginsetal19,Pethetal25,BernNewh24b} for
instance), hence justifying our interest. As a side-benefit, our
proposal also extends the theory presented in \cite{GratToin26c} in
allowing \emph{inexact step normalization strategies}, a clear
advantage in practice for methods like \muon\ where the normalization
is performed using a possibly inexact Newton-Schultz iteration
\cite{Kova70,BjorBowi71,BernNewh24,Vyasetal24,Liuetal25b}. The proposed addition of momentum to
asynchronous steepest-descent methods (the second line for our paper
in Table~\ref{tab:survey}) is also useful to enhance numerical
performance, but it may come with a (mostly theoretical) cost. For the
common momentum definition of using the gradient, assuming that the
step-size is sufficiently small may be necessary to prove convergence,
but the resulting degraded convergence rate may improved by adopting a
suitable strategy for the momentum parameter. Remarkably, no
assumption on the step-size is required for the momentum-less variant,
or for a definition of the momentum using the momentum itself.

The paper is organized as follows. The momentum-less asynchronous
framework is presented in Section~\ref{s:algonomom}, where it is shown
to be a special case of an easier-to-analyze delayed framework, whose
rate of convergence is studied in Section~\ref{s:momentumless}. Two
different variants of momentum are introduced and the resulting rates
of convergence examined in Section~\ref{s:momentum}. Numerical
experiments illustrating the use of asynchronous block-wise adaptive
\muon/\adagrad\ are presented in Section~\ref{s:numerics}.  Finally,
some conclusions and perspectives are discussed in
Section~\ref{s:conclusion} while Appendix~1 and Appendix~2 give the
details of the convergence proofs for the methods of
Sections~\ref{s:algonomom} and \ref{s:momentumless}, respectively.

\textbf{Notations:} In what follows, $\|\cdot\|_E$  denotes the
Euclidean norm. If $\|\cdot\|$ is a norm
on some space $\calS$ endowed with an inner product
$\langle\cdot,\cdot\rangle$, its dual norm $\|\cdot\|_D$ is defined  by
$\|x\|_D = \max_{y\in \calS}\langle y,x \rangle/ \|y\|$. The symbol
$I_n$ denotes the identity matrix of dimension $n$ and $0_n$ the zero
vector of size $n$. If $x$ is a vector, $[x]_i$ denotes its $i$-th
component and $[x]_{\calI}$ the set of the components with indices in
$\calI$. If $\calS$ is a set, $|\calS|$ denotes its cardinal.

\numsection{\laname: a stochastic asynchronous first-order framework}\label{s:algonomom}

The basis of our approach is to assume that the problem variables come
in disjoint \emph{blocks} of homogeneous nature.  In deep learning
applications, this occurs for instance when some variables are layer
weights and other thresholds or biases of activation nodes. The first
may be scalar and the second matrices \cite{Yuetal17,Ginsetal19}. In
line with \cite{GratToin26c}, we assume that there are $L$ blocks of variables, and problem
\req{problem} is therefore cast in the Cartesian product space
\[
\Re^N = \prod_{\ell=1}^L \Re^{n_\ell\times m_\ell} = \prod_{\ell=1}^L\Re^{d_\ell}.
\]
We denote by $\calI_\ell$ the set of indices of the variables in block
$\ell$. For $X\in \Re^n$,  $[X]_{\calI_\ell}$ is then the
subset of variables in $X$ belonging to block $\ell$.
Following \cite{GratToin26c}, we then specify, for each block, a particular
norm $\|\cdot\|_\ell$ and its dual $\|\cdot\|_{\ell,*}$. The inner
product on $\Re^N$ is then defined as
\[
\ip{U,V} = \sum_{\ell=1}^L \ip{U_\ell,V_\ell}_F
\tim{ where }
\ip{A,B}_F = \tr(A^TB)
\]
and the  norm on $\Re^N$ given by
\beqn{big-norm}
\|V\|^2 = \sum_{\ell=1}^L \|V_\ell\|_\ell^2
\eeqn
whose dual norm is
\beqn{big-dual-norm}
\|U\|_*^2 = \sum_{\ell=1}^L \|U_\ell\|_{\ell,*}^2.
\eeqn

Because of the structure of the variables' space, it is natural to
avoid mixing variables of different blocks, and this therefore defines
the geometry of the asynchronous minimization method we are going to
construct. Moreover, as is shown by the practical success of
methods like \tal{Muon} for matrix variables, different minimization
methods may be appropriate for different types of variables. For each
block, we therefore define a collection of methods applicable to
variables in the block\footnote{For instance, \al{AdaNorm}
\cite{WardWuBott19} and \al{Adagrad} \cite{DuchHazaSing11,McMaStre10}
are applicable methods for blocks with scalar 
variables. \al{Shampoo} \cite{GuptKoreSing18} and \al{Muon}
\cite{Jordetal24} are applicable for blocks with 
matrix variables.}.  As was already the case in 
\cite{GratToin26c}, each such \emph{method} for block $\ell$ is specified by
a preconditioner's update operator
$\calU: \Re^{d_\ell}\rightarrow\Re^{d_\ell}$ and a normalization
operator $\calN: \Re^{d_\ell}\rightarrow \Re^{d_\ell}$. The
collection of methods applicable for block $\ell$ is then
given by
$\calM_\ell =\{(\calU_1,\calN_1), \ldots, (\calU_{a_\ell},\calN_{\hat m_\ell})\}$.

While this way to describe first-order adaptively preconditioned
gradient methods may seem abstract, it was shown in \cite{GratToin26c}
that it covers a fairly representative set of popular methods, such as
\tal{AdaNorm}, full and diagonal \tal{Adagrad}, \tal{Shampoo} and
\tal{Muon}\footnote{An adaptive version of \al{Muon}
\cite{ZhanLiuScha25} for block $\ell$ is, for example, given by setting
$\|\cdot\|_\ell$ to the spectral matrix norm, $\|\cdot\|_*$ to its
dual, the nuclear norm, and defining
$\calN_\ell(G) = U V^T$ where the singular value
decomposition of $G$ is given by $U \Sigma V^T$, and
$\calU(G) = (\|G\|_*/\sqrt{d_\ell})I_{d_\ell}$.
} (see \cite[Section~4.5]{GratToin26c} for details).

We now specify the \aname\footnote{{\sf\footnotesize AS}ynchronous {\sf\footnotesize AD}aptive
{\sf\footnotesize PREC}onditioned gradient} algorithmic
\emph{framework} \vpageref{genalgo}, meaning that
its description can be instantiated in a number of different specific
algorithms. In what follows, we use the shorthand ``\saname\ algorithm'' to
refer to any algorithm belonging to the \saname\ framework.
  
\algo{genalgo}{\aname}{
    Given: a starting state $\hX_0$ and constants $\eta,\varsigma>0$. \\
    For $\ell\in\ii{L}$, set $\hPi_\ell=\varsigma I_\ell$ and $[\widehat{F}]_\ell = 1$.\\
    For each processor $p\in\ii{P}$ in parallel, continuously
    \vspace*{1mm}
    \begin{enumerate}
    \item read $X \leftarrow \hX$
      \hfill (block-atomic, inconsistent)
      \vspace*{-2.5mm}
    \item (read $F \leftarrow \widehat{\calF}$)
           \hfill (component-atomic)
      \vspace*{-2.5mm}
    \item select a block $\ell$ such that $[F]_\ell = 1$ and
      an associated method $(\calU_\ell,\calN_\ell)\in\calM_\ell$.
      \vspace*{-2.5mm}
    \item (update $[\widehat{F}]_\ell = 0$)
           \hfill (component-atomic)
      \vspace*{-2.5mm}
    \item draw $\xi_\ell$ and set $\tG_\ell = [\nabla_X f(X,\xi_\ell)]_{\calI_\ell}$
      \vspace*{-2.5mm}
    \item (read $\Pi_\ell^-\leftarrow \hPi_\ell$)
          \hfill (block-atomic)
      \vspace*{-2.5mm}
    \item $\Pi_\ell =\Pi_\ell^-+\calU_\ell(\tG_\ell)^2$,
      \vspace*{-2.5mm}
    \item $Z_\ell = \Pi_\ell^{-1/2}\tG_\ell$,
      \vspace*{-2.5mm}
    \item (update $\hPi_\ell \leftarrow \Pi_\ell$)
          \hfill (block-atomic)
      \vspace*{-3mm}
    \item update $[\hX]_{\calI_\ell} \leftarrow [X]_{\calI_\ell}-\eta\,\|Z_\ell\|_{*,\ell}\,\calN_\ell(Z_\ell)$
      \hfill (block-atomic)
      \vspace*{-3mm}
    \item (update $[\widehat{F}]_\ell = 1$)
          \hfill (component-atomic)
    \end{enumerate}
}

\noindent
Some comments on this algorithm are useful at this stage.
\begin{enumerate}
\item We have stated the framework in a form as general as
  possible. It is however important to observe that \emph{significant
  simplifications occur if}, as is often practical, \emph{the computing architecture allocates to
  each processor a set of blocks disjoint from that of other
  processors}. In that setting, the preconditioner $\hPi_\ell$ may be
  stored locally on the processor in charge of block $\ell$ and Steps
  6 and 9 are unnecessary.  Moreover, this design also ensures that
  not two processors can work on the same block at the same time, so
  that Steps 2, 3 , 4 and 11 can be replaced by a simpler Step~3 where
  the block selection is restricted to those assigned to the current
  processor. Thus all steps in brackets can be ignored and Step~3
  simplified, thereby significantly reducing interprocessor
  communication costs and resulting in a much leaner operation.
\vspace*{-2mm}
\item Pushing asynchronicity to the component- (rather than block-)
  level is difficult for $\Pi_\ell$ if one wishes to allow for non-diagonal
  preconditioners (as in full \adagrad\ or \shampoo), because
  inconsistently read symmetric matrices may fail to be positive-definite.
\vspace*{-2mm}
\item No writing conflict can occur in Steps 9 and 10 because we have
  assumed a totally separable geometry, with no overlap between
  blocks.
\vspace*{-2mm}
\item The computation of the block's gradient $\tG_{k(\ell_t)}$ in Step~4
  may or may not require the computation of the complete gradient
  $\nabla f(X_{k(\ell_t)},\xi_{k(\ell_t)})$, depending on the particular
  form of the objective function and the computer architecture.
  
\vspace*{-2mm}
\item The general framework of \aname\ is reminiscent of the
  block-Jacobi \cite{BertTsit89,FromSzyl00} setting for solving linear
  systems of equations. 
\end{enumerate}

\noindent
Our analysis of \aname's global rate of convergence is based on the
view that, from an analytical standpoint, it  can
be viewed as a synchronous algorithmic framework with delays.
 To establish this interpretation, we define a \emph{global iterate}
  $\hX_t$ $(t=0,\ldots)$ as the state of the shared memory $\hX$ at
 each moment $t$ where $\hX$ is modified by a block-atomic update.
  Because of the read operation of Step~1 is atomic, we
  have that $X = \hX_t$ and, after Step~5, $\tG_\ell=[\nabla_X
    f(\hX_t,\xi_\ell)]_{\calI_{\ell_t}}$. If $\ell_t$ is the index
  then selected at Step~3, the algorithm's loop on block $\ell_t$
  continues, starting from $(\hX_t,[\nabla_X f(\hX_t,\xi_{k(\ell_t)})]_{\calI_{\ell_t}})$. 
  Now consider the interval $t$ to $t+1$ and all steps being computed
  in parallel in this interval.  Ignore all such steps whose
  computation does not terminate at $t+1$, and focus on the (unique,
  with probability one) block $q_t$ which terminates at $t+1$.
  The computation of this step was started at
  some time  $t-\tau_{t,q_t}$ (where the nonnegative integer $\tau_{t,q_t}$
  is called a \emph{delay}) using $\hX_{t-\tau_{t,q_t}}$ and the
  approximate block gradient
  $[\nabla_X f(\hX_{t-\tau_{t,q_t}},\xi_{t-\tau_{t,q_t}})]_{\calI_{q_t}}$.
  Moreover, because the selection process at Step~3 prevents two different
  processors to work on block $q_t$ at the same time, we have that
  $[\hX_t]_{\calI_{q_t}}=[\hX_{t-\tau_{t,q_t}}]_{\calI_{q_t}}$. 
  We may then (formally) consider that this step on block $q_t$ was
  started at time $t$   from $\hX_t$, but using the   ``delayed
  approximate block gradient''  
  $[\nabla_Xf(\hX_{t-\tau_{t,q_t}},\xi_{k(q_{t-\tau_{t,q_t}})})]_{\calI_{q_t}}$. Thus the
  sequence of overlapping step computations for different blocks is
  equivalent to a sequence where, in each interval from $t$ to $t+1$, only the block
  steps terminating at the end of the interval are considered and are
  deemed to have started at $t$ using a delayed approximate block gradient.
  This is exactly how the \asname\ algorithm described
  \vpageref{algo:dadprec} proceeds, at each iteration selecting a
  subset of blocks for which a complete step is computed using a
  approximate, \emph{possibly delayed}, block gradient. We may thus see the sequence of iterates
  $\hX_t$ generated by an \aname\ algorithm as being instead
  generated by a \asname\ algorithm, thereby identifying $\hX_t$ for
  \aname\ with $X_t$ for \asname. We also identify
  $\tG_{\ell_t}$, $\Pi_{\ell_t}$ and $Z_{\ell_t}$ in the former with $\tG_{t,\ell}$,
  $\Pi_{t,\ell}$ and $Z_{t,\ell}$ in the latter.
  
\algo{algo:dadprec}{\asname}{
  Given: a starting point $X_0$ and constants $\eta,\varsigma>0$. Set
  $\Pi_{-1,\ell}=\varsigma I_\ell$ for $\ell\in\ii{L}$.\\
  For $t = 0, 1, \ldots$ 
  \begin{enumerate}
  \item Draw $\xi_t$ and compute $\tG_t$.
      \vspace*{-2.5mm}
  \item Select a nonempty ``active'' block set $\calA_t\subseteq \ii{L}$.
      \vspace*{-2.5mm}
  \item For $\ell\in \calA_t$, select
    $(\calU_{t,\ell},\calN_{t,\ell})\in\calM_\ell$ and compute 
      \vspace*{-2.5mm}
    \begin{eqnarray}
    \Pi_{t,\ell} & = &\Pi_{t-1,\ell}+\calU_{t,\ell}(\tG_{t,\ell})^2,\label{Pi-def}\\
    Z_{t,\ell}   & = & \Pi_{t,\ell}^{-1/2}\tG_{t,\ell},\label{z-def}\\
    X_{t+1,\ell} & = & X_{t,\ell} - \eta\,\|Z_{t,\ell}\|_{*,\ell}\,\calN_{t,\ell}(Z_{t,\ell}).\label{xkp1}
    \end{eqnarray}
      \vspace*{-8mm}
  \item For $\ell\not\in\calA_t$, (formally) set
      \vspace*{-2.5mm}
    \begin{eqnarray}
    \Pi_{t,\ell} & = &\Pi_{t-1,\ell},\label{Pi-def2}\\
    Z_{t,\ell}   & = & 0_{n_\ell\times m_\ell}\label{z-def2}\\
    X_{t+1,\ell} & = & X_{t,\ell}\label{xkp12}
    \end{eqnarray}
      \vspace*{-8mm}
   \end{enumerate}
}

Note that Step~4 of \asname\  is purely formal, because that it does not
involve any computation. Note also that $\tG_{t,\ell}$ in \asname\ is a
fairly general approximation of the true block gradient, as it may
include delays and other errors, such as resulting from sampling.

The reason for introducing the \asname\ framework is that its
convergence analysis can be derived, with moderate effort, from that
of the proposal in \cite{GratToin26c}. We describe the necessary
assumptions and the results obtained (for \asname\ and thus for
\aname) in the next section, and postpone to the
Appendix the detailed description of the
modifications of the analysis of \cite{GratToin26c}.

\vspace*{2mm}
\noindent
\fbox{\fbox{\parbox{0.965\linewidth}{
All assumptions and results will thus be formulated using the notation
of the \asname\ framework, but may directly be interpreted in the
context of \aname.
}}}
\vspace*{2mm}

As is clear from the algorithms' statements, this analysis  does not
include momentum. Variants including momentum 
will be considered in Section~\ref{s:momentum}.

\numsection{Convergence of the momentum-less \lasname}\label{s:momentumless}

Our analysis of the \asname\ framework starts by formulating
standard conditions on the problem \req{problem}.

\begin{assumption}[Boundedness]
\label{ass:bounded-op}
There exists a constant $\flow$ such that $F(X) \ge \flow$ for all
$X\in\Re^N$.
\end{assumption}

\noindent
Defining $G(X) = \nabla_X F(X)$, we then require the following
smoothness assumption.

\begin{assumption}[Smoothness]
\label{ass:smooth-op}
The objective function $f$ is continuously differentiable and has a Lipschitz continuous gradient,
that is there exists a constant $L_G\ge 0$ such that, for all $X,Y\in\Re^N$,
$
\|G(X)-G(Y)\|_*\le L_G\|X-Y\|.
$
\end{assumption}

\noindent
We now introduce conditions that define the class of methods
$(\calU_\ell,\calN_\ell)$ for which we develop our theory.  These are
defined for a given block $\ell\in\ii{L}$ and depends on the choice
of dual norm $\|\cdot\|_{\ell,*} $ for this block. Again, we stress
that \textit{these abstract conditions do hold for several popular
  methods} for specific choices of the methods
$(\calU,\calN)\in\calM_\ell$, including \tal{AdaNorm}, full and
diagonal \tal{Adagrad}, 
as well as adaptive variants of \tal{Shampoo} and
\tal{Muon} \cite[Section~4.5]{GratToin26c}. 

\begin{assumption}[Structural identities]
  \label{ass:identities}
For each $t\geq 0$ and $\ell\in\calA_t$, we have that, for all
$(\calU,\calN) \in \calM_\ell$,
\beqn{ineq1}
\|Z_{t,\ell}\|_{*,\ell}\,\ip{G_{t,\ell},\calN(Z_{t,\ell})}_F
= \tr\!\big(\Pi_{t,\ell}^{-1/2}\calU(\tG_{t,\ell})^2\big),
\eeqn
and
\beqn{ineq2}
\|Z_{t,\ell}\|_{*,\ell}^2=\tr\!\big(\Pi_{t,\ell}^{-1}\calU(\tG_{t,\ell})^2\big).
\eeqn
\end{assumption}

\begin{assumption}[Gradient-preconditioner compatibility]
\label{ass:opttransfer}
There exists a constant $\kappa_\circ > 0$ such
that, for all $\calU$ in $(\calU,\calN)\in\calM_\ell$ and all $G$,
\begin{equation}
\label{eq:generic}
\|G\|_{*,\ell}^2 \le \kappa_\circ \, \tr\!\big(\,\calU(G)^2\big).
\end{equation}
\end{assumption}

\noindent
We also need to consider stochastic conditions on the quality of the
gradient oracle. 

\begin{assumption}[Unbiased oracle]
\label{ass:unbiased}
The block gradient oracle is unbiased, that is
\beqn{no-bias}
\Econd{t}{\tG_{t,\ell}} = G_{t,\ell}
\eeqn
for all $t \ge 0$ and all $\ell\in \calA_t$.
\end{assumption}

\begin{assumption}[Cumulative variance]\label{ass:variance}
There exists a constant $\omega\ge0$ and a sequence
$\{\nu_t\}_{k\ge0}$, such that, for all $t\ge 0$,
\beqn{var-cond}
\sum_{j=0}^t \sum_{\ell\in\calA_j}\E{\|\tG_{j,\ell}-G_{j,\ell}\|_*^2}
\le \nu_t^2 + \omega^2\sum_{j=0}^t\sum_{\ell\in\calA_j}\E{\|Z_{j,\ell}\|_{*,\ell}^2}.
\eeqn
\end{assumption}

\noindent
Observe that Assumption~\ref{ass:unbiased} is made at the block level,
and is weaker than assuming the more standard condition that
$\Econd{t}{\tG_t} = G_t= G(X_t)$. Similarly, Assumption~\ref{ass:variance} is
also made at the block level. Moreover \req{var-cond} requires a bound on
the \textit{cumulative} variance of all block gradient oracles over all past
iterates, an approach also more general than assuming conditional variance
at every iteration. In particular, it allows large variance at early
iterations provided later iterations compensate.  Trade-offs between
different realizations and/or different blocKs are also
theoretically possible. 

We finally impose a minimal condition on the asynchronous nature of
the algorithm.

\begin{assumption}[Bounded delays]\label{ass:delays}
There exists an integer $\tau \ge 0$ such that $\tau_{t,\ell} \le\tau$
for all $t\ge0$ and all $\ell\in\ii{L}$.
\end{assumption}

\noindent
For the purpose of analysis, we also define, for $t\geq 0$ and $\ell\in\ii{L}$,
\[
\calJ_{t,\ell} = \{j\in\iiz{t} \mid \ell \in \calA_j\},
\]
the set of indices of iterations at which block $\ell$ is effectively
updated, and immediately note that the summations $\sum_{j=0}^t\sum_{\ell\in\calA_j}$ and
$\sum_{\ell=1}^L\sum_{j\in\calJ_{t,\ell}}$ are interchangeable (for
instance in \req{var-cond}).

\vspace*{2mm}
A general theorem stating the global rate of convergence of any
\asname\ (and thus \aname) algorithm may now be stated under these assumptions.

\lthm{thm:dadprec-convergence}{
Suppose that the \asname\ algorithm is applied to problem
\req{problem} and that Assumptions~\ref{ass:bounded-op} to
\ref{ass:delays} hold.
Then, for all $t \ge 0$,
\beqn{thetruerate}
\vspace*{-2mm}
\min_{j\in\iiz{t}} \E{\|G_j\|_*}
\le \frac{1}{t+1}\sum_{j=0}^t \E{\|G_j\|_*}
\le \frac{\kappa_a + \kappa_b\,\sqrt{N\log(\Theta_t)}+\kappa_c\,\Theta_t}{\sqrt{t+1}}
\vspace*{-2mm}
\eeqn
with
\[
\kappa_a = L_G\eta\sqrt{\tau L \kappa_0},
\ms
\kappa_b = L_G\eta\sqrt{2\tau L}
\tim{ and }
\kappa_c = \kappa_\circ\tau\sqrt{L},
\]
\beqn{Thetak-def}
\Theta_t
\eqdef
\max\left[\kappa_\Theta, 12\sqrt{N}\,
\nu_t\,\sqrt{\max\left[1,\log\left(12\sqrt{N}\,\nu_t\right)\right]}\right]
\vspace*{-2mm}
\eeqn
with
\vspace*{-2mm}
\[
\kappa_\Theta =\max\left[ e^{\max[1,\frac{\kappa_0}{2N}]},\,\E{f(X_0)}-f_{\rm low} + \eta\,\varsigma\,N,
24N\left(\omega+\frac{L_G}{\eta}\right)
\log\left(24N\left(\omega+\frac{L_G}{\eta}\right)\right)\right]
\]
and $\kappa_0=-\sum_{\ell=1}^L d_\ell\log(d_\ell) -
N\log(\varsigma)$.
}
\proof{See Appendix~1.}

\noindent
Note that, since we prove (in Appendix) that each preconditioner
$\Pi_{t,\ell}$, is bounded by the (in expectation) very slowly growing
$\Theta_t$ (see \req{Thetak-def}), and since $Z_t$ is the
preconditioned approximate gradient, condition \req{var-cond} is akin
to a (undelayed) cumulative affine variance condition of the form
\[
\sum_{j=0}^s\sum_{\ell=1}^L \Econd{j}{\|\tG_{j,\ell}-G_{j,\ell}\|^2}
\le \nu_t^2 +\omega^2\sum_{j=0}^s\sum_{\ell=1}^L\Econd{j}{\|G_{j,\ell}\|_{*,\ell}^2},
\]
whose iteration-wise variant has already been used in analysis of
first-order methods (see \cite{Fawetal22,WangZhanMaChen23} or, for the
stronger `` affine-$^*$ '' version, \cite{AttiKore23}). In particular, the ``strong growth''
assumption motivated by over-parametrized problems and used in
\cite{WangZhanMaChen23} to derive an improved 
convergence rate for \adagrad\ is essentially subsumed
(at the cumulative level) for ``preconditioned cumulative strong growth''
(that is if $\nu_t=0$ is assumed in \req{var-cond}).

\noindent
As stated, the rate of convergence of \asname\  is dominated by the
term in $\kappa_c\Theta_t/\sqrt{t+1}$ in \req{thetruerate}, where
$\Theta_t$ depends on $\nu_t$, the bound on the cumulative variance of
the block gradient oracles, which may itself depend on $t$. The next
corollary describes what can be said if one makes a more specific
assumption on the oracle (block) variance at each iteration.

\llcor{therate2}{
Suppose that the \asname\ algorithm is applied to problem
\req{problem}  Assumptions~\ref{ass:bounded-op} to \ref{ass:unbiased}
and \ref{ass:delays} hold. Suppose also that, for all $t \ge 0$ and $\ell\in\calA_t$,
\beqn{betaisktogamma}
\Econd{t}{\|\tG_{t,\ell}-G_{t,\ell}\|_*^2}
\le \frac{\sigma_\ell^2}{|\calJ_{t,\ell}|^\alpha}+\omega^2\Econd{t}{\|Z_{t,\ell}\|_*^2}
\eeqn
for some $\sigma_\ell\ge 0$, $\ell\in\calA_t$ and $\alpha,\omega >0$.
Then, if $\psi_t = \max_{\ell\in\ii{L}}|\calJ_{t,\ell}|$,
\beqn{therate2-psi}
\frac{1}{t+1}\sum_{j=0}^t \E{\|G_j\|_*}=
\left\{\begin{array}{ll}
\calO\left(\bigfrac{\psi_t^{1-\alpha}\sqrt{\log(\psi_t)}}{\sqrt{t+1}}\right)
&\tim{if } \alpha<1,\\
\calO\left(\bigfrac{\sqrt{\log(\psi_t)\log(\log(t+1))}}{\sqrt{t+1}}\right)
&\tim{if } \alpha=1,\\
\calO\left(\bigfrac{1}{\sqrt{t+1}}\right)
&\tim{if } \alpha>1.
\end{array}\right.
\eeqn
}

\proof{See Appendix~1.}

\noindent
From the \aname\ point of view, $|\calJ_{t,\ell}|$ corresponds to $k(\ell_t)$, and
therefore counts the number of updates to block $t$, while $\psi_t$ is
their maximum taken on all blocks.
We have that $\psi_tc\le t+1$ in general, but if, at each iteration $t$,
$|\calA_t| \ll L$ (a not unusual situation), then
$\psi_t \ll t+1$, suggesting an acceleration of the convergence rate
compared with the undelayed algorithm (where $\psi_k = t+1$).

Observe that the continuity of the bound expressed in Theorem~\ref{thm:dadprec-convergence}  as
a function of $\nu_t$ is lost in the statement of
Corollary~\ref{therate2}, because the
constants hidden in the $\calO(\cdot)$ notation depend on $\alpha$. In
particular, this formulation does not support taking
the limit for $\alpha$ tending to one.
Observe also that we could weaken \req{betaisktogamma} by requiring that
\beqn{alt-var-cond}
\Econd{t}{\|\tG_{t,\ell}-G_{t,\ell}\|^2} \le \frac{\sigma_\ell^2}{t^\alpha}+\omega^2\|Z_{t-1,\ell}\|_*^2
\eeqn
for $\ell\in\calA_t$ with the convention that $Z_{-1,\ell}=0$,
since, obviously,
\[
\sum_{\ell=1}^L\sum_{j=0}^{t-1}\Econd{j}{\|Z_{j,\ell}\|_*^2}
\le \sum_{\ell=1}^L\sum_{j=0}^t\Econd{j}{\|Z_{j,\ell}\|_*^2}.
\]
The advantage of \req{alt-var-cond} over
\req{betaisktogamma} is that the right-hand-side of this new condition
is now measurable at iteration $t$.

As was noted in \cite{GratToin26c},
although the rates of convergence obtained in Corollary~\ref{therate2} under the
bias and variance Assumptions~\ref{ass:unbiased} and \ref{ass:variance} are reasonable, they
do not quite match, for high-variance regimes, the best obtained so far for
(momentum-less, synchronous) AdaGrad and AdaNorm \cite{WangZhanMaChen23}.  However
they recover (in order) the best
$\calO(\sqrt{\log(k+1)\log(\log(k+1))/(k+1)})$ rate in the preconditioned cumulative strong growth
context\footnote{That is not only in the case where $\sigma_\ell = 0$ for each $\ell$,
but, more generally, in the case where the ``inaccuracy
budget'' $\nu_t^2$  is finite.}. Importantly, they do so for a large
class of asynchronous algorithms.

\numsection{Convergence of \lasname\ with momentum}\label{s:momentum}

Adding momentum to first-order methods has often been instrumental in
improving numerical performance\footnote{Although, as far as the
authors are aware, this has not so far been translated in improved convergence theory.}.
It is therefore of interest to consider asynchronous versions of such
methods.  We investigate two versions of this idea in this section,
which share the common framework \anameM\ stated below for some
given sequences $\{\mu_k\}$ of momentum parameters with
$0\le \mu_k\le \mu_{\max} < 1$ for all $k\ge 0$,
with $k$ indexing the number of updates to block $\ell$.

\algo{genalgo-mom}{\tal{\anameM}}{
  Given: a starting state $\hX_0$ and constants $\eta,\varsigma>0$. \\
  For $\ell \in\ii{L}$, set $\hPi_{-1,\ell}=\varsigma I_\ell$.\\
  For each processor $p\in\ii{P}$ in parallel, continuously
    \begin{enumerate}
    \item read $X \leftarrow \hX_t$ \hfill 
      (block-atomic, inconsistent)
    \vspace*{-2.5mm}
    \item (read $F \leftarrow \widehat{F}$)
           \hfill (component-atomic)
    \vspace*{-2.5mm}
    \item select a block $\ell$ such that $[F]_\ell = 1$ and
      an associated method $(\calU_\ell,\calN_\ell)\in\calM_\ell$.
      \vspace*{-2.5mm}
    \item (update $[\widehat{F}]_\ell = 0$)
           \hfill (component-atomic)
      \vspace*{-2.5mm}
    \item (read $\Pi_\ell^-\leftarrow \hPi_\ell$) \hfill (block-atomic)
    \vspace*{-2.5mm}
    \item (read $M_\ell^-\leftarrow \hM_\ell$) \hfill   (block-atomic)
    \vspace*{-2.5mm}
    \item draw $\xi_\ell$ and set $\tG_\ell = [\nabla_X f(X,\xi_\ell)]_{\calI_\ell}$
    \vspace*{-2.5mm}
    \item (read $k \leftarrow [\widehat{K}]_\ell$)
           \hfill (component-atomic)
      \vspace*{-2.5mm}
    \item \fbox{compute $M_\ell$ and $\Pi_\ell$, using $\calU_\ell$
      and $k$.}
    \vspace*{-2.5mm}
    \item $Z_\ell  = \Pi_\ell^{-1/2}M_\ell$,
    \vspace*{-2.5mm}
    \item (update $\hPi_\ell \leftarrow \Pi_\ell$)       \hfill (block-atomic)
    \vspace*{-2.5mm}
    \item (update $\hM_\ell \leftarrow M_\ell$)       \hfill (block-atomic)
    \vspace*{-2.5mm}
    \item (update $[\widehat{F}]_\ell = 1$)
           \hfill (component-atomic)
      \vspace*{-2.5mm}
    \item (update $[\widehat{K}]_\ell = k+1$)
           \hfill (component-atomic)
      \vspace*{-2.5mm}
    \item update $[\hX]_{\calI_\ell}
       \leftarrow [X]_{\calI_\ell}-\eta\,\|Z_\ell\|_{*,\ell}\,\calN_\ell(Z_\ell)$\hfill (block-atomic)
    \end{enumerate}
}
\noindent
The two variants are specified by detailing Step~9 either as
\[
\hspace*{-1mm}\left.\begin{array}{lcl}
\Pi_\ell &= &\Pi_\ell^-+\calU_\ell(\tG_\ell)^2,\\
M_\ell &=  & \left\{\begin{array}{ll}
      \mu_k M_\ell^-+(1-\mu_k) \tG_\ell &\tim{if } k>0,\\
      \tG_\ell &\tim{if } k=0,
\end{array}\right.
\end{array}
\hspace*{9mm} \right\}(\anameMG)
\]
or as
\[
\hspace*{-1mm}\left.\begin{array}{lcl}
M_\ell & = & \left\{\begin{array}{ll}
      \mu_kM_\ell^-+(1-\mu_k) \tG_\ell &\tim{if } k>0,\\
      \tG_\ell &\tim{if } k=0 \end{array}\right.\\
\Pi_\ell & =& \Pi_\ell^-+\calU_\ell(M_\ell)^2.
\end{array}
\hspace*{9mm} \right\}(\anameMM)
\]
Note that, in both cases, the momentum $\widehat{M}_\ell$ and the
index $[\widehat{K}]_\ell$ must in general be read from and updated
in shared memory, unless, again, specific blocks are affected to specific
processors, in which case they can be maintained in local memory. \emph{All
bracketed steps may thus be omitted in this context.}

Fortunately, the convergence theory can be adapted to cover both
variants. This is achieved again by considering \asnameMM\ and
\asnameMG, the obvious extensions of \asname\ corresponding to
\anameMM\ and \anameMG, respectively,
where \req{Pi-def} and \req{z-def} are now replaced by

\noindent
\begin{tabular}{c|c}
\begin{minipage}{0.49\linewidth}
\vspace*{-2mm}
\begin{eqnarray*}
M_{t,\ell}   & =&  \mu_{t,\ell} M_{\pi(t-1,\ell),\ell}+(1-\mu_{t,\ell})\tG_{t,\ell},\label{MmomMM}\\
\Pi_{t,\ell} & =&  \Pi_{t-1,\ell}+ \calU_{t,\ell}(M_{t,\ell})^2,\label{GamomMM}\\
Z_{t,\ell}   & =&  \Pi_{t,\ell}^{-1/2}M_{t,\ell}\label{ZmomMM}
\end{eqnarray*}
\hspace*{15mm}for \asnameMM, 
\end{minipage}
&
\begin{minipage}{0.49\linewidth}
\vspace*{-2mm}
\begin{eqnarray*}
\Pi_{t,\ell} & =&  \Pi_{t-1,\ell}+ \calU_{t,\ell}(\tG_{t,\ell})^2,\label{GamomMG}\\
M_{t,\ell}   & =&  \mu_{t,\ell} M_{\pi(t-1,\ell),\ell}+(1-\mu_{t,\ell})\tG_{t,\ell},\label{MmomMG}\\
Z_{t,\ell}   & =&  \Pi_{t,\ell}^{-1/2}M_{t,\ell}\label{ZmomMG}
\end{eqnarray*}
\hspace*{15mm}for \asnameMG.
\end{minipage}
\end{tabular}\\

\noindent
Because the update formula for $Z_{t,\ell}$ above now
involves the momentum $M_{t,\ell}$ instead of $\tG_{t,\ell}$, we
reformulate Assumption~\ref{ass:identities} to reflect this
change\footnote{The reformulation obviously still holds for \adanorm,
\adagrad, \shampoo\ and \muon\ \cite{GratToin26c}.} as follows.

\begin{assumption}[Structural identities with momentum]\label{ass:mom-identities}
For each $t\ge0$ and $\ell\in \calA_t$, we have that, for all
$(\calU,\calN) \in \calM_\ell$,
\vspace*{-2mm}
\beqn{ineq1-mom}
\|Z_{t,\ell}\|_{*,\ell}\,\ip{M_{t,\ell},\calN(Z_{t,\ell})}_F
= \tr\!\big(\Pi_{t,\ell}^{-1/2}\calU(M_{t,\ell})^2\big),
\eeqn
and
\vspace*{-2mm}
\beqn{ineq2-mom}
\|Z_{k(\ell)}\|_{*,\ell}^2=\tr\!\big(\Pi_{k(\ell)}^{-1}\calU(M_{k(\ell)})^2\big).
\eeqn
\end{assumption}

\noindent
Armed with this reformulated assumption, it is now possible to establish
that the expected global rate of convergence of a \asnameMM\ 
algorithm is essentially identical to that of the version without
momentum.

\lthm{thm:momMM-convergence}{
Suppose that an \asnameMM\ algorithm is applied to problem
\req{problem} and that Assumptions~\ref{ass:bounded-op}, \ref{ass:smooth-op}, 
and \ref{ass:opttransfer} to \ref{ass:mom-identities} hold. Then the
conclusions of Theorem~\ref{thm:dadprec-convergence} and
Corollary~\ref{therate2} do hold.
}

\proof{See Appendix~2.}

\noindent
We note at this point that our momentum definition and convergence
analysis for the \asnameMM\ algorithm do not
make the assumption that the stepsize parameter $\eta$ is sufficiently
small, in contrast with previous proofs of convergence where results
assume that $\eta$ is small enough, in particular smaller than a multiple
of the (usually unknown) Lipschitz constant $L_G$
\cite{GuoYinJinYang21,HongLin24,XiaoHuLiuToh24}.
Establishing convergence results for the \asnameMG\ variant
unfortunately requires additional assumptions.  In particular, the
stepsize $\eta$ has to be chosen small enough. This alternative theory
hinges on the fact that
$\calU(\tG_{k,\ell})$ can be viewed as a perturbation of
$\calU(M_{k,\ell})$ , and therefore that the structural relations of
Assumption~\ref{ass:mom-identities} are themselves perturbed when
the \asnameMG\ algorithm is used.
Fortunately, it remains possible to bound these perturbations by a mix
of variance-related and second-order terms, the first of which
potentially affecting the resulting global convergence
rate. The final outcome is given by the following theorem and
corollary.

\lthm{thm:convergence-alt}{
Suppose that a \asnameMG\ algorithm is applied to problem \req{problem}.
Suppose that Assumptions~\ref{ass:bounded-op}, \ref{ass:smooth-op},
\ref{ass:opttransfer}, \ref{ass:unbiased}, \ref{ass:delays} and \ref{ass:mom-identities}
hold. Suppose also that there exists constants
$\kappa_\Box,\kappa_\diamond>0$ such that 
 \vspace*{-2mm}
 \beqn{ass:sub-add}
  \calU(U+V)^2 \preceq \kappa_\Box \calU(U)^2 + \kappa_\Box \calU(V)^2
  \tim{ and }
  \tr\!\Big(\calU(U)^2\Big) \le \kappa_\diamond \|U\|_{*,\ell}^2
  \eeqn
\vspace*{-2mm}
  for all $t\ge0$ and $\ell\in\calA_t$ and all $U,V\in\calM_\ell$,
  and that 
  \beqn{small-eta}
  \tim{either }
  \eta \le \frac{1-\mu_{\max}}{L_G}
  \sqrt{\frac{\varsigma}{6\kappa_\Box\kappa_\diamond\tau L}}
  \tim{ ~or~ }
    \sum_{j=0}^\infty \|Z_j\|_*^2  \le \kappa_Z
\vspace*{-3mm}
  \eeqn
for some $\kappa_Z\ge 0$. Then
\vspace*{-2mm}
\beqn{thetruerate-alt}
\min_{j\in\iiz{t}} \E{\|G_j\|_*}
\le \frac{1}{t+1}\sum_{j=0}^t \E{\|G_j\|_*}
\le \frac{\kappa_a + \kappa_b\,\sqrt{N\log(\Theta_t)}+\kappa_c\,\Theta_t}{\sqrt{t+1}}
\eeqn
with
\vspace*{-2mm}
\[
\kappa_a = L_G\eta\sqrt{\tau L \kappa_0},
\ms
\kappa_b = L_G\eta\sqrt{2\tau L},
\ms
\kappa_c = \kappa_\circ\tau\sqrt{L},
\]
and
\beqn{Thetak-def-alt}
\Theta_t = \max\left[\,\kappa_\Theta, \frac{3\kappa_{\nu\nu}\theta_t^2}{\eta}, T_t \right],
\vspace*{-2mm}
\eeqn
where
\vspace*{-2mm}
\[
\kappa_\Theta =
\max\left[\,e^{\max\left[1,\frac{\kappa_0}{2N}\right]},
  \,\frac{3\kap{gap}}{\eta},
  \, 24N\kappa_\Delta\left(\omega^2+\frac{L_G}{\eta}\right)
  \log\left(24N\kappa_\Delta\left(\omega^2+\frac{L_G}{\eta}\right)\right)
  \right],
\]
\vspace*{-2mm}
\beqn{var-cond-alt}
\theta_t^2 = \sum_{\ell=1}^L\sum_{j\in\calJ_{t,\ell}}\max[\mu_{\pi(j-1,\ell),\ell},\mu_{j,\ell}]^2\,\E{\|\tG_{j,\ell}-G_{j,\ell}\|_{*,\ell}^2},
\eeqn
\vspace*{-2mm}
\[
T_t = 12\sqrt{N}\,\kappa_{\nu\Delta}\,
\theta_t\,\sqrt{\max\left[1,\log\left(12\sqrt{N}\,\kappa_{\nu\Delta}\,\theta_t\right)\right]},
\]
$\kap{gap} =\E{f(X_0)}-f_{\rm low} + \eta\,\varsigma\,N$,
$\kappa_\circ$ is defined in Assumption~\ref{ass:opttransfer},
$\kappa_0$ in Theorem~\ref{thm:dadprec-convergence}, and
$\kap{gap}$, $\kappa_{\nu\nu}$, $\kappa_{\nu\Delta}$ and 
$\kappa_\Delta$ are specified in \req{kaps-def1}--\req{kaps-def2}.
}

\proof{See Appendix~2.}

\noindent
As it turns out, assuming \req{ass:sub-add} is not
restrictive\footnote{\adanorm, \adagrad, \shampoo\ and
\muon\ continue to be covered, as verified in \cite{GratToin26c}.},
but supposing \req{small-eta} is definitely less desirable because the
value of the Lipschitz constant $L_G$  or that of $\kappa_Z$ (if
it exists) is usually unknown.  It is
however common practice in the literature
\cite{GuoYinJinYang21,HongLin24,XiaoHuLiuToh24}.
Note the term $\kappa_{\nu\nu}\nu_t^2$ in the middle term of
the expression of $\Theta_t$ in \req{Thetak-def-alt}, which is the
first significant difference between
\req{Thetak-def-alt} and \req{Thetak-def} and
induces a modified convergence rate. The second difference is the presence of the factor
$\max[\mu_{\pi(j-1,\ell),\ell},\mu_{j,\ell}]^2$ in the definition of the
cumulative variance, which has the opposite effect of potentially
improving the convergence rate.  This is expressed by the following
corollary, where $|\calJ_{t,\ell}|$ is the number of updates to block
$\ell$ up to iteration $t$.

\llcor{therate2-alt}{
Suppose that a \asnameMG\ algorithm is applied to problem \req{problem}.
Suppose that Assumptions~\ref{ass:bounded-op}, \ref{ass:smooth-op},
\ref{ass:unbiased}, \ref{ass:opttransfer}, \ref{ass:delays} and
\ref{ass:mom-identities} hold.
Suppose also that \req{ass:sub-add} and \req{small-eta}
hold, that
\beqn{var-cond-alt2}
\Econd{t}{\|\tG_{t,\ell}-G_{t,\ell}\|_*^2} \le
\frac{\sigma_\ell^2}{|\calJ_{t,\ell}|^\alpha}+\omega^2 \Econd{t}{\|Z_{t,\ell}\|_{*,\ell}^2}
\eeqn
for some $\alpha>0$ and all $k \ge 0$ and $\ell\in\ii{L}$, and that,
for each $\ell \in\calA_t$,
\beqn{mutell}
\mu_{t,\ell} \in \left[0, \frac{\mu_{\max}}{|\calJ_{t,\ell}|^\beta}\right]
\eeqn
for some $\mu_{\max}<1$ and some $\beta\ge 0$.
Then, if $\psi_t=\max_{\ell\in\ii{L}}|\calJ_{t,\ell}|$,
\beqn{therateMG}
\frac{1}{k+1}\sum_{j=0}^t \E{\|G_j\|_*}=
\left\{\begin{array}{ll}
\calO\left(\bigfrac{\psi_t^{1-\alpha-2\beta}}{\sqrt{t+1}}\right)
&\tim{if } \alpha+2\beta<1,\\
\calO\left(\bigfrac{\log(\psi_t)}{\sqrt{t+1}}\right)
&\tim{if } \alpha+2\beta=1,\\
\calO\left(\bigfrac{1}{\sqrt{t+1}}\right)
&\tim{if } \alpha+2\beta>1.
\end{array}\right.
\eeqn
}
\proof{See Appendix~2.}

\noindent
Again, be aware that the constants hidden in the $\calO(\cdot)$ depend
on $\alpha$ and $\beta$, in particular preventing taking limits for
$\alpha$ tending to one.  The rate of convergence now depends on the
value of $\alpha+2\beta$, indicating how \emph{acting on the momentum
parameter can partly alleviate the effect of high gradient oracle's
variance}. Because the convergence rate is determined by $\theta_t$,
choosing a small momentum parameter in effect reduces the propagation
of larger errors in the gradient oracle across iterations. In
particular, setting $\mu_{t,\ell}$ to a multiple of
$|\calJ_{t,\ell}|^{-\quarter\alpha}$ recovers the rate of the
momentum-less variant also for the high-variance regime
($\alpha<1$). If however $\mu_{t,\ell}$ is kept constant (or bounded
away from zero), that is if $\beta=0$, then \req{var-cond-alt2} is
stronger than \req{betaisktogamma} and the rate of convergence for
$\alpha<1$ is now $\calO\left((k+1)^{\half-\alpha}\right)$ instead of
$\calO\left((k+1)^{-\sfrac{\alpha}{2}}\right)$, requiring in
particular that $\alpha > \half$.

\numsection{Approximate normalization}

We have assumed, in our theory, that normalization of the step is
exact, in the sense that $\|\calN_{t,\ell}(Z)\|_\ell=1$ for all $t$, all
$\ell\in\calA_t$ and all $Z\in\Re^{n_\ell \times m_\ell}$. This choice
has been made for simplicity, but the reader can readily convince
himself/herself that it is not necessary, and that it is sufficient
to assume the existence of constants $0<\kappa_{1,\calN} \le 1 \le
\kappa_{2,\calN}$ such that 
$\|\calN_{t,\ell}(Z)\|_\ell\in [\kappa_{1,\calN}, \kappa_{2,\calN}]$ for
all $t$, $\ell\in\calA_t$ and all $Z\in\Re^{n_\ell \times m_\ell}$.  The
constants $\kappa_{1,\calN}$ and $\kappa_{2,\calN}$ then percolate
though all proofs, complicating the expression of relevant factors
even more, but without affecting the order of global convergence.
This observation is particularly useful for methods involving matrix
blocks, such as \shampoo\ or \muon, where step normalization is
typically performed by a possibly inexact Newton-Schulz iteration
\cite{Kova70,BjorBowi71,BernNewh24,Vyasetal24,Liuetal25b}.
  
\numsection{Numerical experiments}\label{s:numerics}

\subsection{Experimental objectives}\label{sec:objectives}

The experiments in this section illustrate the practical behavior
of the asynchronous block-parallel adaptive preconditioning method
(\aname) introduced in the previous sections.
The convergence analysis in Section~\ref{s:momentumless} establishes
that the asynchronous block-coordinate iteration converges under a
bounded-delay condition, provided the block preconditioners
capture sufficient local curvature to control the inter-block
coupling.
The experiments are designed to verify that these theoretical
predictions hold on concrete machine learning problems, and to
quantify the practical trade-offs involved.

The OFFO (Objective-Function-Free Optimization) framework
of~\cite{GratToin26c} provides the algorithmic foundation:
the parameter vector is decomposed into $L$~blocks, each
equipped with its own adaptive preconditioner, and the method
requires only gradient evaluations---no function value
computations.
In the synchronous regime, this reduces to a preconditioned
block-Jacobi iteration; in the asynchronous regime, each block
is updated independently using a possibly stale snapshot of the
other blocks.
Classical results on asynchronous chaotic
relaxation~\cite{ChazMira69} and asynchronous
iterations~\cite{BertTsit89,FromSzyl00}
predict convergence when the spectral radius of the
``asynchronous iteration matrix'' is less than one---a condition
that is favored by weak inter-block coupling and strong local
preconditioning.
In practice, the throughput gain from asynchronous execution
depends on the computational heterogeneity between blocks.
When all blocks have equal per-iteration cost, asynchronous
execution offers no advantage over the synchronized variant
(the barrier imposes no idle time).
When block costs differ, the synchronization barrier forces fast
blocks to idle while waiting for the slowest block, and the
throughput gain from removing the barrier is bounded by an
Amdahl-type ratio: the cost of the slowest worker relative to
the average.
The experiments are designed to explore this relationship across
a range of heterogeneity levels.
More specifically, the experiments address the following questions:
\begin{enumerate}
\vspace*{-2mm}
\item Does asynchronous block-parallel optimization preserve
convergence quality (training loss, gradient norm, generalization
metric) relative to its synchronous counterpart, as predicted
by the bounded-delay convergence theory?
\vspace*{-2mm}
\item What throughput gain does asynchronous execution provide, and
how does this gain relate to the measured computational heterogeneity
between blocks (Amdahl-type bound)?
\vspace*{-2mm}
\item How does exponential moving average (EMA) momentum interact
with the stale reads inherent to asynchronous execution?
\end{enumerate}

\noindent
The benchmarks are chosen to cover a range of architectural
patterns (dense MLPs, sparse embedding models, hybrid architectures)
and heterogeneity levels, so that the interplay between block
structure, computational cost imbalance, and asynchronous scheduling
can be studied in a controlled setting.
The goal is \emph{not} to achieve state-of-the-art predictive
performance, but to verify the theoretical predictions and to
identify the regimes in which asynchronous block parallelism
is most beneficial. We consider three algorithmic variants.
All three apply the same per-block preconditioned updates and differ
only in scheduling and synchronization.
\begin{description}
\item[Synchronous single-gradient (\syncsingle): ]
A single mini-batch is sampled and the stochastic gradient
$G_k = \nabla F_{\mathcal{B}_k}(x_k)$
is computed once. All $L$~blocks are updated from the same snapshot,
using  Steps~8 and 10 of the \aname\ framework.
This variant minimizes gradient computation cost but assumes
that a single gradient evaluation can be shared across all block
solvers.
\vspace*{-2mm}
\item[Synchronous multi-gradient (\syncmulti): ]
Each of the $L$~workers independently computes the full stochastic
gradient on the same shared batch and snapshot, then extracts its own
block component.
Workers are synchronized by barriers: no block update is applied
until every worker has completed its gradient computation.
The mathematical iterate is identical to that of, 
\syncsingle, using Steps~8 and 10 of \aname, but the wall-clock cost per
iteration includes the overhead of $L$~gradient evaluations and
barrier synchronization.
\vspace*{-2mm}
\item[Asynchronous block-owner (\async): ]
The synchronization barriers are removed.
Each block~$X_\ell$ is permanently assigned to a dedicated worker.
All workers share access to the full parameter vector
$\hX = (\hX_1, \dots, \hX_L)$.
Worker~$\ell$ repeatedly:
\begin{enumerate}
\item[(i)] reads the full vector $\hX$ by cloning each block $\hX_j$
  under block~$j$'s lock, ensuring that all parameters within a
  given block come from the same publication epoch
  (intra-block consistency); different blocks may come from
  different epochs (inter-block asynchrony);
\vspace*{-2mm}
\item[(ii)] samples a local mini-batch $\mathcal{B}_{t,\ell}$,
  computes the full stochastic gradient
  $\nabla F_{\mathcal{B}_{t,\ell}}(\hX_k)$, and extracts
  the block component
  $G_{t,\ell} = [\nabla F_{\mathcal{B}_{t,\ell}}(\hX_k)]_\ell$;
\vspace*{-2mm}
\item[(iii)] applies the preconditioned update to $\hX_\ell$ and
  publishes the result atomically under block~$\ell$'s lock.
\end{enumerate}
This is the asynchronous block-Jacobi
model~\cite{BertTsit89,FromSzyl00} used by \aname.
A bounded-delay throttle pauses any worker whose publication counter
exceeds the slowest worker's by more than $\tau_{\max} = 512$.
\end{description}

\subsection{Test problems and architectures}\label{sec:problems}

Table~\ref{tab:architecture} summarizes the five benchmarks.
Each model is decomposed so that one or two blocks use the
matrix-valued adaptive \muon\  preconditioner
(Newton--Schulz  polar approximation
\cite{Jordetal24} for SVHN, SVD for the others)
while the remaining blocks use diagonal \adagrad.

\begin{table}[htbp]
\centering
\footnotesize
\begin{tabular}{l l r r r l}
\toprule
Dataset & Architecture & $n_{\mathrm{train}}$ & $p$ & $L$
  & Muon block(s) \\
\midrule
FashionMNIST
  & MLP $784 \to 4096 \to 10$
  & 60{,}000 & 3{,}256{,}330 & 4
  & $W_1 \in \mathbb{R}^{4096\times 784}$ \\[3pt]
MoE-FMNIST
  & 4 experts (see text)
  & 60{,}000 & 3{,}887{,}950 & 6
  & \begin{tabular}[t]{@{}l@{}}$W_{1,1} \in \mathbb{R}^{512\times 2048}$,\\
    $W_{2,1} \in \mathbb{R}^{512\times 1024}$\end{tabular} \\[3pt]
CovType
  & MLP $54 \to 512 \to 256 \to 7$
  & 160{,}000 & 161{,}287 & 4
  &  $W_2 \in \mathbb{R}^{512\times 256}$ \\[3pt]
MovieLens 100K
  & NeuralMF, $d{=}128$, $h{=}512$
  & 80{,}000 & 470{,}722 & 5
  & $W_1\in \mathbb{R}^{512\times 256}$ \\[3pt]
Criteo
  & Hash-embed + MLP $429 \to 256 \to 1$
  & 160{,}000 & 634{,}625 & 3
  & $W_1 \in \mathbb{R}^{429\times 256}$ \\[3pt]
SVHN
  & MLP $3072 \to 4096 \to 512 \to 10$
  & 73{,}257 & 14{,}689{,}802 & 5
  & \begin{tabular}[t]{@{}l@{}}$W_1 \in \mathbb{R}^{4096\times 3072}$,\\
    $W_2 \in \mathbb{R}^{512\times 4096}$\end{tabular} \\
\bottomrule
\end{tabular}
\caption{Test problems and block decompositions.}
\label{tab:architecture}
\end{table}

\begin{description}
\item[FashionMNIST~\cite{XiaoRasuVoll17}: ]
A one-hidden-layer MLP with $3.26 \times 10^6$ parameters
decomposed into 4~blocks:
$W_1$ (\muon), $b_1$, $W_2$, $b_2$ (\adagrad).

\item[MoE-FMNIST~\cite{XiaoRasuVoll17}: ]
  A Mixture-of-Experts
  architecture applied to flattened 28×2828×28 Fashion-MNIST images,
  with a backbone MLP ($784 \rightarrow 512784\rightarrow 512$), a top-2/4 softmax router, four
  experts of varying width ($512 \rightarrow h \rightarrow 512$) with
  $h\in\{2048,1024,256,64\}$), and an output head
  ($512 \rightarrow 10512 \rightarrow 10$),
  decomposed into 6 blocks: $W_{1,1}$ and $W_{2,1}$
  (\muon, Newton--Schulz, 5 iterations), expert 3 and expert 4 full
  parameters, backbone and expert 1 and 2 biases and second-layer
  weights, router and output head (all \adagrad).
\item[CovType~\cite{BlacDean99}: ]
A two-hidden-layer MLP with $1.61 \times 10^5$ parameters
(200{,}000 samples, 80/20 split), decomposed into 4~blocks:
$(W_1,b_1)$, $W_2$ (\muon), $b_2$, $(W_3,b_3)$ (\adagrad).

\item[MovieLens 100K~\cite{HarpKons15}: ]
A NeuralMF model with $4.71 \times 10^5$ parameters
(100{,}000 ratings, 943~users, 1{,}682~items),
decomposed into 5~blocks:
$P$, $Q$, $(b_u, b_i)$ (sparse row-wise AdaGrad),
$W_1$ (\muon), $(c_1, W_2, c_2)$ (\adagrad).

\item[Criteo~\cite{Crit14}: ]
A click-through-rate model with $6.35 \times 10^5$ parameters,
decomposed into 3~blocks:
embedding table (AdaGrad), $W_1$ (\muon),
$(b_1, W_2, b_2)$ (\adagrad).

\item[SVHN~\cite{Netzetal11}: ]
A two-hidden-layer MLP with $14.7 \times 10^6$ parameters
applied to flattened $32\times 32\times 3$ images,
decomposed into 5~blocks:
$W_1$ and $W_2$ (\muon, Newton--Schulz, 5 iterations),
$b_1$, $b_2$, $(W_3, b_3)$ (\adagrad).
This benchmark features two Muon blocks of different
sizes ($12.6 \times 10^6$ and $2.1 \times 10^6$ entries), creating
a hierarchy of block costs.
\end{description}

\subsubsection{Measured block heterogeneity}\label{sec:heterogeneity}

The raw block cost ratio ($t_{\max}/t_{\min}$) exceeds $90$ on
all benchmarks, but the effective heterogeneity is considerably lower.
Each worker's per-iteration cost has two components:
the gradient computation $t_{\mathrm{grad}}$ (identical for all workers,
since each computes the full gradient $\nabla F$) and the block update
$t_{\mathrm{block},\ell}$. When $t_{\mathrm{grad}}$ is large relative
to $t_{\mathrm{block},\ell}$, all workers run at similar speeds
regardless of block update costs. Table~\ref{tab:heterogeneity}
reports per-block computational costs measured on an Apple~M3~Pro
processor (single-threaded, median over 50~repetitions).
The \emph{effective heterogeneity ratio} is defined as
$(t_{\mathrm{grad}} + t_{\max}) / (t_{\mathrm{grad}} + t_{\min})$.

\begin{table}[htbp]
\centering
\footnotesize
\begin{tabular}{l r r r r r}
\toprule
Dataset     & Gradient & Slowest block & Fastest block & $t_{\max}/t_{\min}$ & Eff.\ ratio \\
\midrule
FashionMNIST & $5.6$  & $105.2$ (\muon) & $0.004$  & $25{,}000$  & $19.6$ \\
MoE-FMNIST   & $37.6$ & $451.0$ (\muon) & $0.035$  & $12{,}733$  & $5.7$ \\
CovType      & $0.8$  & $0.42$  (\muon) & $0.004$  & $96$        & $1.5$ \\
MovieLens    & $1.0$  & $5.61$  (\muon) & $0.011$  & $526$       & $6.5$ \\
Criteo       & $1.8$  & $4.83$  (\muon) & $0.012$  & $419$       & $3.7$ \\
SVHN         & $23.4$ & $580.2$ (\muon) & $0.004$  & $134{,}000$ & $25.8$ \\
\bottomrule
\end{tabular}
\caption{Per-block computational cost (ms) and effective heterogeneity.}
\label{tab:heterogeneity}
\end{table}

On FashionMNIST, the \muon\ polar factorization ($105$~ms) far exceeds
the gradient ($5.6$~ms), so the effective ratio remains high ($19.6$).
On CovType, the gradient ($0.8$~ms) exceeds the
slowest block update ($0.42$~ms), reducing the effective ratio to
$1.5\times$.
SVHN exhibits the highest effective ratio ($25.8$) thanks to
the Newton--Schulz iteration on $W_1 \in \mathbb{R}^{4096 \times 3072}$
($580$~ms) relative to the gradient ($23$~ms).

\subsection{Implementation and experimental protocol}\label{sec:protocol}

All experiments use PyTorch on a shared-memory CPU architecture
(Apple~M3~Pro, 12~cores) with \texttt{torch.set\_num\_threads(1)}.
The CPython GIL prevents byte-code level parallelism, but
PyTorch C++ operations (matrix multiplications, SVD, tensor cloning)
release the GIL, allowing concurrent execution across cores.
The \async\ mode uses per-block locking:
each block is read and written under its own lock, ensuring
intra-block consistency while allowing inter-block asynchrony.
A fixed wall-clock budget of 30~seconds is used for problems CovType,
MovieLens and Criteo, a budget of 60~seconds for Fashion-MNIST,
and MoE-FMNIST and a budget of 120~seconds for SVHN.
All runs use 8~seeds ($1234, \dots, 1241$), and identical initial parameters per seed.
Throughput is reported in model-equivalent updates per second:
$\mathrm{throughput} = N_{\mathrm{block}} / (L \cdot T)$.

\subsection{Results without momentum}\label{sec:results}

Tables~\ref{tab:nomom-conv} and \ref{tab:nomom-gen} present the
results obtained with the momentum-less 
\aname\ in the setting just described.
We start by reporting, for all benchmarks the full-dataset
training loss and the Euclidean norm of the full training gradient
$\|\nabla F(X)\|$ as convergence diagnostics.

\begin{table}[htbp]
 \centering
 \footnotesize
 \begin{tabular}{|l|ccc|ccc|}
   \hline
   & \multicolumn{3}{c|}{Final training loss gradient $\|\nabla F(X)\|$ }
   &\multicolumn{3}{c|}{Final training loss value $F(X)$}\\*[0.8ex]
   Dataset & \syncsingle  & \syncmulti & \sasync & \syncsingle  & \syncmulti & \sasync \\
   \hline
   FashionMNIST     &    0.421     &    0.343   & \textbf{0.277} & 0.321  & 0.321 & \textbf{0.308}\\
   MoE-FMNIST &\textbf{0.632} &    0.826   &   0.744  & 0.309  &   0.314 & \textbf{0.279}\\
   CovType          &    0.568     & \textbf{0.482} &   0.485  & 0.243  &   0.260 & \textbf{0.240}\\
   MovieLens        &    0.024     &    0.027   & \textbf{0.021} & \textbf{0.294}  &   0.340 &   0.353\\
   Criteo           &    0.044     &    0.053   & \textbf{0.033} & \textbf{0.076} &   0.112 &   0.088\\
   SVHN             &    1.004     & \textbf{0.979} &   1.180  & 0.840  &   0.833 & \textbf{0.819}\\ 
   \hline
 \end{tabular}
 \caption{Convergence diagnostics for momentum-less variants}
 \label{tab:nomom-conv}
\end{table}

\sasync\ obtains the best gradient (in norm) for Criteo, FashionMNIST and
MovieLens, while there is little difference between \syncsingle\ and
\sasync\ for CovType. The gradient norm is higher for SVHN and MoE-FMNIST
(due to delayed gradients). The best loss value is obtained by
\sasync\ for 4/6 benchmarks. \syncsingle\ remains better for MovieLens and
Criteo because it performs only one gradient evaluation per iteration.
   
We now turn to primary generalization metrics, depending on the benchmark.
For classification problems (FashionMNIST, CovType, SVHN), we
report the \emph{test accuracy} (fraction of correctly classified
test samples).
For MovieLens, we report the \emph{root mean square error} (RMSE),
defined as
$\mathrm{RMSE} = \sqrt{\frac{1}{n}\sum_{i=1}^n (\hat r_i - r_i)^2}$,
where $\hat r_i$ and $r_i$ are the predicted and true ratings.
For Criteo, we report the \emph{area under the ROC curve} (AUC),
which measures the probability that a randomly chosen positive
example (clicked ad) is ranked higher than a randomly chosen
negative example by the model's predicted score; AUC $= 0.5$
corresponds to random ranking and AUC $= 1$ to perfect ranking.
Generalization metrics are reported in Table~\ref{tab:nomom-gen},
along with the throughput of \sasync\ and the speedup of
\sasync\ compared with \syncmulti.

\begin{table}[htbp]
 \centering
 \footnotesize
 \begin{tabular}{|l|l|ccc|c|c|}
   \hline
   Dataset & metric & \syncsingle  & \syncmulti & \sasync & Throughput & Speedup\\
   \hline
   FashionMNIST     & acc.(\%) & 87.00 & 86.94 & \textbf{87.34} & 10   & 1.43 \\
   MoE-FMNIST & acc.(\%) & 86.82 & 86.52 & \textbf{87.39} & 42   & 1.27 \\
   CovType          & acc.(\%) & 89.49 & 88.82 & \textbf{89.56} & 105  & 1.27 \\
   MovieLens        & RMSE     & 0.930 & 0.919 & \textbf{0.916} & 89   & 1.14 \\
   Criteo           & AUC      & 0.715 & \textbf{0.722} & 0.717 & 89   & 1.17 \\
   SVHN             & acc.(\%) & 69.44 & 69.28 & \textbf{69.65} & 1    & 1.33 \\
   \hline
 \end{tabular}
 \caption{Generalization metrics, throughput and speed-up for momentum-less variants}
 \label{tab:nomom-gen}
\end{table}

\sasync\ is best on all benchmarks except Criteo.

Globally, Tables~\ref{tab:nomom-conv} and
\ref{tab:nomom-gen} suggest that stale (delayed) gradients do not degrade the
quality of the training and that the additional iterations made
possible by asynchrony translate into better generalization. 

\subsection{Results with momentum}\label{sec:momentum}

The approaches yielding \asnameMM/\anameMM\ and \asnameMG/\anameMG\ not only
differ in theory, but their practical requirements are also
significantly different, in particular for the \shampoo and
\muon\ case. Indeed, if the SVD of $M_\ell$ is given by
$M_\ell = U_\ell \Sigma_\ell V_\ell^T$, the \muon\ recursions\footnote{$\Pi_\ell$ is a
scalar in this case.} are written, in this case, as\\
\noindent
\begin{tabular}{c|c}
\begin{minipage}{0.49\linewidth}
\vspace*{-2mm}
\begin{eqnarray*}
M_\ell   & = & \mu_k M_\ell^-+(1-\mu_k) \tG_\ell \\
\Pi_\ell & = &\Pi_\ell^-+\frac{\|M_\ell\|_{*,\ell}^2}{d_\ell},\\
X_\ell^+ & = & X_\ell -\eta
\frac{\|M_\ell\|_{*,\ell}}{\sqrt{\Pi_\ell}} U_\ell V_\ell^T
\end{eqnarray*}
\hspace*{15mm} for \muon.{\sf MM},
\end{minipage}
&
\begin{minipage}{0.49\linewidth}
\vspace*{-2mm}
\begin{eqnarray*}
\Pi_\ell & = &\Pi_\ell^-+\frac{\|\tG_\ell\|_{*,\ell}^2}{d_\ell},\\
M_\ell   & = & \mu_k M_\ell^-+(1-\mu_k) \tG_\ell \\
X_\ell^+ & = & X_\ell -\eta
\frac{\|M_\ell\|_{*,\ell}}{\sqrt{\Pi_\ell}}  U_\ell V_\ell^T
\end{eqnarray*}
\hspace*{15mm} for \muon.{\sf MG}.
\end{minipage}
\end{tabular}\\*[1ex]
The main difference is that \muon.{\sf MM} only requires
$\|M_\ell\|_{*,\ell}$, which can be computed with a single polar
decomposition (using SVD or Newton-Schulz iterations), while
\muon.{\sf MG} also requires $\|\tG_\ell\|_{*,\ell}$. This potentially
requires another decomposition, which can significantly slow
down the algorithm.  It is therefore of interest to consider faster
techniques for computing $\|\tG_\ell\|_{*,\ell}$.  We have considered
both the stochastic Lanczos quadrature \cite{UbarChenSaad17} and a warm-started power
iteration for this task. As it turns out, the first, although
theoretically suitable, appears to be too slow for practical use. By
contrast, the second, which can only provide a deterministic lower
bound, nevertheless works remarkably well. The details of this
technique are given in Appendix~3.

When applied to \aname, these considerations lead us to a family of momentum variants,
of which we have retained six, whose characteristics are given in
Table~\ref{tab:mom-vars}. In this table, $\beta$ is the rate of momentum decay appearing in
\req{mutell} (with $\mu_{\max}=0.9$) and the last column indicates whether a warm-started
power iteration is used to approximate  $\|\tG_\ell\|_{*,\ell}$ in the
\muon\ updates.

\begin{table}[htbp]
  \centering \footnotesize
  \begin{tabular}{|l|l|c|c|}
      \hline
      Variant & type & $\beta$ & approx. $\|\tG_\ell\|_{*,\ell}$ \\
      \hline
          {\sf MMconst} & MM & 0 & no \\
          {\sf MGconst} & MG & 0 & no \\
          {\sf MGconstA}& MG & 0 & yes \\
          {\sf MMdecay} & MM & 0.5 & no \\
          {\sf MGdecay} & MG & 0.5 & no \\
          {\sf MGdecayA}& MG & 0.5 & yes \\
      \hline
  \end{tabular}
  \caption{Asynchronous momentum variants and their characteristics}
  \label{tab:mom-vars}
\end{table}

To keep the paper as concise as possible, we only report in
Table~\ref{tab:mom-vars-grad} the values of the final gradients (in
norm) for these six variants, and refer the reader to Appendix~3 for
more results.

\begin{table}[htbp]
\centering \footnotesize
\begin{tabular}{|l|c|c|c|c|c|c|c|}
\hline
Dataset & No Mom. & {\sf MMconst} & {\sf MGconst} & {\sf MGconstA} &{\sf MMdecay} & {\sf MGdecay} & {\sf MGdecayA}\\
\hline
Fashion-MNIST  & 0.277 & 0.551 & 0.500 & 0.526 & 0.255 & \textbf{0.246} & 0.312 \\
MoE-FMNIST     & 0.744 & 0.858 & \textbf{0.656} & 1.212 & 0.663 & 0.884 & 0.881 \\
CovType        & 0.485 & \textbf{0.246} & 0.279 & 0.265 & 0.345 & 0.386 & 0.755 \\
MovieLens      & 0.021 & 0.021 & 0.036 & 0.032 & \textbf{0.015} & 0.019 & 0.025 \\
Criteo         & 0.033 & 0.036 & \textbf{0.008} & 0.018 & 0.046 & 0.035 & 0.028 \\
SVHN           & 1.180 & 1.430 & 0.988 & 1.138 & 1.103 & \textbf{0.880} & 1.131 \\
\hline
\end{tabular}
\caption{Final gradient norm for momentum variants}
\label{tab:mom-vars-grad}
\end{table}

\noindent
As can be seen from this table, no specific momentum variant appears
to dominate, but the momentum-less variant (first column) is never the best.

\subsection{Discussion}\label{sec:discussion}

Globally, our results suggest the following conclusions.

\begin{enumerate}
\item \emph{Asynchronous parallelism preserves and improves
convergence quality.} \asname\ is the
  best momentum-less variant for 5/6 benchmarks, and is also best in
  many case where momentum is used.
\item\emph{Heterogeneity drives the speedup.}
The per-block locking protocol---guaranteeing intra-block
consistency while allowing inter-block asynchrony---matches the
classical asynchronous block-Jacobi model~\cite{BertTsit89,FromSzyl00}.
The effective heterogeneity ratio, which accounts for the shared
gradient computation cost, is a better predictor of the asynchronous
speedup than the raw block cost ratio.
\vspace*{-2mm}
\item The \emph{cost of momentum is critical for matrix updates} like \muon.
This makes the {\sf MM} approach attractive from this point of view, because its lower cost per
iteration allows more iterations in the time budget, resulting in
gains in accuracy between +1.8 and +3.6 points compared with the {\sf
  MG} approach.
\vspace*{-2mm}
\item The estimation of the nuclear norm of $\tG_\ell$ using the
  \emph{warm-started power iteration is an efficient proxy}, and does produce
  the best results for some problems.
\vspace*{-2mm}
\item As expected, the effect of decaying momentum (i.e.\ $\beta>0$ in
  \req{mutell}) is problem-dependent, because of its direct
  interaction with the variance of the problem's oracle.
\vspace*{-2mm}
\item Although globally promising, our tests did not allow us to select a clear winner among
  the studied variants.
\end{enumerate}

\noindent
It should however be stressed that all our experiments were
conducted assuming the ``full gradient evaluation model'', that is the
case where $G_{t,\ell} = [\nabla
  F_{\mathcal{B}_{t,\ell}}(\hX_k)]_\ell$ cannot be computed on its
own, but requires the evaluation of $\nabla
F_{\mathcal{B}_{t,\ell}}(\hX_k)$. Should the evaluation of
$G_{t,\ell}$ be possible at a lower cost\footnote{Among other possibilities, we think here of
structured networks such as mix of experts \cite{HampWaib92}, DeepONets
\cite{LuJinKarn19} or multi-frequency networks \cite{GratMercRiccToin24}. The
approach of delayed gradients suggested by
\cite{ZhuaWangLiuZhanLin19}, in some sense close to \asname,
may also be worth investigating.},
the heterogeneity ratios would then be dramatically increased, which is
likely to give the asynchronous option used in \aname\ a substantial
advantage.

\numsection{Conclusions and perspectives}\label{s:conclusion}

We have presented a new class of asynchronous adaptive first-order
methods including asynchronous variants of \adanorm, \adagrad,
\shampoo\ and \muon, as well as variants thereof using momentum and/or
inexact step normalization. We have shown that, under reasonable
assumptions, all methods in the class have an global convergence rate
(essentially) of order $\calO(1/\sqrt{t})$ in expectation.

We also described numerical experiments with these methods, providing
motivation for asynchronous adaptive optimization where different
blocks use different geometries, different blocks have drastically
different computational costs, and synchronization barriers become
prohibitively expensive.  Such situations naturally appear in modern
large-scale machine learning systems mixing diagonal adaptive methods,
Our results, although limited, provide empirical support for
the claim that asynchronous adaptive block-coordinate optimization
methods may be relevant in heterogeneous large-scale machine learning
systems.


\appendix
\appnumsection{Appendix~1: Detailed analysis for momentum-less
  \lasname}

The proof of Theorem~\ref{thm:dadprec-convergence} follows the broad
lines of \cite{GratToin26c}. In order to keep our argument as compact
as possible, we refer, in what follows, to that reference
for the proofs of results directly extracted from it\footnote{Notation
has changed: in \cite{GratToin26c}, the iteration index was $k$, the normalization operator
$S_\ell(\cdot)$ and the preconditioner update operator $\calL_{k,\ell}(\cdot)$.}.

Our first step is to analyze the (expected) descent at
iteration $t$.

\llem{cond-desc}{
Suppose that Assumption~\ref{ass:smooth-op} holds. Then
\beqn{descent-k}
\begin{aligned}
\Econd{k}{f(X_{t+1})}
\le & f(X_t)-\eta\sum_{\ell\in\calA_t}\Econd{t}{\|Z_{t,\ell}\|_{*,\ell}\ip{\tG_{t,\ell},\calN_{t,\ell}(Z_{t,\ell})}_F}\\
&+\eta\sum_{\ell\in\calA_t}\Econd{t}{\|Z_{t,\ell}\|_{*,\ell}\big|\ip{\tG_{t,\ell}-G_{t,\ell},\calN_{t,\ell}(Z_{t,\ell})}_F\big|}
+\frac{L_G\eta^2}{2}\sum_{\ell\in\calA_t}\Econd{t}{\|Z_{t,\ell}\|_{*,\ell}^2}.
\end{aligned}
\eeqn
}
\proof{See \cite[Lemma~3.1]{GratToin26c}.}

\noindent
The following trace inequalities may then be proved.

\llem{lem:sqrt-op}{Suppose that a \asnameMG\ algorithm is applied to problem \req{problem}.
Consider the \asname\ algorithm.
We have that, for all $k\ge 0$,
\beqn{sqrt-pot}
\sum_{\ell=1}^L \tr(\Pi_{t,\ell}^{1/2})-\sum_{\ell=1}^L\tr(\Pi_{-1,\ell}^{1/2})
\le \sum_{\ell=1}^L\sum_{j\in\calJ_{k,\ell}}\tr\!\left(\Pi_{j,\ell}^{-1/2}\calU_{j,\ell}(\tG_{j,\ell})^2\right)
\vspace*{-3mm}
\eeqn
\vspace*{-1mm}
and
\beqn{log-pot}
\sum_{\ell=1}^L\sum_{j\in\calJ_{t,\ell}}\tr\!\left(\Pi_{j,\ell}^{-1}\calU_{j,\ell}(\tG_{j,\ell})^2\right)
\le\sum_{\ell=1}^L\tr\!\Big(\log(\Pi_{t,\ell})\Big)-\sum_{\ell=1}^L\tr\!\Big(\log(\Pi_{-1,\ell})\Big).
\eeqn
}

\proof{See \cite[Lemmas 3.2, 3.3 and 3.4]{GratToin26c}.  In the last
  of these lemmas, the double sums $\sum_{j=0}^k\sum_{\ell=1}^L$ are
  replaced by $\sum_{\ell=1}^L\sum_{j\in\calJ_{t,\ell}}$.
}

\noindent
Because not every block is necessarily updated at iteration $t$ in the
\asname\ algorithm, we now need to re-prove the suitable
variant of the classical telescoping argument. 

\lthm{th:master-op}{
Suppose that Assumption~\ref{ass:bounded-op}, \ref{ass:smooth-op},
\ref{ass:identities} and \ref{ass:variance} hold.
Define
\beqn{Delta-def}
\Delta_t
\eqdef \E{\sum_{\ell=1}^L \tr(\log\Pi_{k,\ell})-\sum_{\ell=1}^L \tr(\log\Pi_{-1,\ell})}.
\eeqn
Then, for every $k\ge0$,
\beqn{master}
\eta\,\E{\sum_{\ell=1}^L \tr(\Pi_{t,\ell}^{1/2})}
\le \kap{gap}+\eta\,\nu_t\,\sqrt{\Delta_t}
+\left(\eta\omega+\frac{L_G\eta^2}{2}\right)\,\Delta_t,
\eeqn
where $\kap{gap} \eqdef \E{f(X_0)}-f_{\rm low} + \eta\,\varsigma\,N$.
}

\proof{
Taking the full expectation in the conditional descent inequality
\req{descent-k} and summing for $j=0$ to $t$ gives
\beqn{descent-more}
\begin{aligned}
\eta\,\sum_{\ell=1}^L\sum_{j\in\calJ_{t,\ell}}&\E{\|Z_{j,\ell}\|_{*,\ell}\,\ip{\tG_{j,\ell},\calN_{j,\ell}(Z_{j,\ell})}_F}
\le \E{f(X_0)}-f_{\rm low}\\
&\hspace*{-5mm}+\eta\,\sum_{\ell=1}^L\sum_{j\in\calJ_{t,\ell}}\E{\|Z_{j,\ell}\|_{*,\ell}\,\big|\ip{\tG_{j,\ell}-G_{j,\ell},\calN_{j,\ell}\ell(Z_{j,\ell})}_F\big|}
+\frac{L_G\eta^2}{2}\sum_{\ell=1}^L\sum_{j\in\calJ_{t,\ell}}\E{\|Z_{j,\ell}\|_{*,\ell}^2\|\calN_{j,\ell}(Z_{j,\ell})\|_\ell^2}.
\end{aligned}
\eeqn
Using successively \req{ineq1} and \req{sqrt-pot}, we obtain that
\beqn{lin-term}
\begin{aligned}
\eta \,\sum_{\ell=1}^L\sum_{j\in\calJ_{t,\ell}}\E{\|Z_{j,\ell}\|_{*,\ell}\,\ip{\tG_{j,\ell},\calN_{j,\ell}(Z_{j,\ell})}_F}
&=\eta\,\sum_{\ell=1}^L\sum_{j\in\calJ_{t,\ell}}\E{\tr\!\bigl(\Pi_{j,\ell}^{-1/2}\calU_{j,\ell}(\tG_{j,\ell})^2\bigr)}\\
&\ge \eta\,\E{\sum_{l=1}^L\tr(\Pi_{t,\ell}^{1/2})
     -\sum_{\ell=1}^L\tr(\Pi_{-1,\ell}^{1/2})}.
\end{aligned}
\eeqn
Now observe that the Cauchy-Schwarz inequality and the definition of
the normalization operator $\calN_{j,\ell}$ give that
\[
\big|
\ip{\tG_{j,\ell}-G_{j,\ell},\calN_\ell(z_{j,\ell})}_F
\big|
\le
\|\tG_{j,\ell}-G_{j,\ell}\|_{*,\ell}\,
\|\calN_\ell(Z_{j,\ell})\|_\ell
= \|\tG_{j,\ell}-G_{j,\ell}\|_{*,\ell}.
\]
Therefore, for $j\in\calJ_{t,\ell}$,
\[
\E{\|Z_{j,\ell}\|_{*,\ell}\,\big|\ip{\tG_{j,\ell}-G_{j,\ell},\calN_{j,\ell}(Z_{j,\ell})}_F\big|}
\le \sqrt{\E{\|\tG_{j,\ell}-G_{j,\ell}\|_{*,\ell}^2}}\,\sqrt{\E{\|Z_{j,\ell}\|_{*,\ell}^2}}.
\]
The Cauchy-Schwarz inequality also implies that, for
any vectors $a$ and $b$ with nonnegative components,
\beqn{CSsqrt}
\sum_j\sqrt{a_j}\sqrt{b_j}
\le \|\sqrt{a}\|_E\,\|\sqrt{b}\|_E
= \sqrt{\sum_j a_j} \sqrt{\sum_j b_j}.
\eeqn
Hence, summing over $\ell\in\calA_j$ and using \req{big-dual-norm} and \req{z-def2}, we obtain that 
\[
\begin{aligned}
\sum_{\ell\in\calA_j}\E{\|Z_{j,\ell}\|_{*,\ell}\,\big|\ip{\tG_{j,\ell}-G_{j,\ell},\calN_{j,\ell}(Z_{j,\ell})}_F\big|}
&\le\sqrt{\sum_{\ell\in\calA_j}\E{\|\tG_{j,\ell}-G_{j,\ell}\|_*^2}}\,\sqrt{\sum_{\ell\in\calA_j}\E{\|Z_{j,\ell}\|_{*,\ell}^2}}.
\end{aligned}
\]
Summing now over $j\in\iiz{t}$, noting that
$\sum_{j=0}^t\sum_{\ell\in\calA_j}= \sum_{\ell=1}^L\sum_{j\in\calJ_{t,\ell}}$ and using \req{CSsqrt} again, we
deduce
that
\[
\begin{aligned}
\eta\,\sum_{\ell=1}^L\sum_{j\in\calJ_{t,\ell}}&\E{\|Z_{j,\ell}\|_{*,\ell}\,\big|\ip{\tG_{j,\ell}-G_{j,\ell},\calN_{j,\ell}(z_{j,\ell})}_F\big|}\\
&\le\eta\,\sqrt{\sum_{\ell=1}^L\sum_{j\in\calJ_{t,\ell}}\E{\|\tG_{j,\ell}-G_{j,\ell}\|_*^2}}\,
          \sqrt{\sum_{\ell=1}^L\sum_{j=0}^t\E{\|Z_{j,\ell}\|_{*,\ell}^2}}\\
& \le \eta \nu_t
          \sqrt{\sum_{j=0}^t\sum_{\ell=1}^L\E{\|Z_{j,\ell}\|_{*,\ell}^2}}
          + \eta\omega
          \sum_{j=0}^t\sum_{\ell=1}^L\E{\|Z_{j,\ell}\|_{*,\ell}^2}\\
\end{aligned}          
\]
where we used \req{var-cond} to obtain the last inequality.
But, by \req{ineq2}, \req{log-pot} and \req{Delta-def},
\beqn{Delta-low}
\sum_{j=0}^t\sum_{\ell=1}^L\E{\|Z_{j,\ell}\|_{*,\ell}^2}
= \sum_{\ell=1}^L\sum_{j\in\calJ_{t,\ell}}\E{\|Z_{j,\ell}\|_{*,\ell}^2}
=\sum_{j=0}^t\sum_{\ell=1}^L\E{\tr\!\bigl(\Pi_{j,\ell}^{-1}\calU_{j,\ell}(\tG_{j,\ell})^2\bigr)}
\le \Delta_t,
\eeqn
so that
\beqn{noise-term}
\eta\sum_{j=0}^t\sum_{\ell=1}^L\E{\|Z_{j,\ell}\|_{*,\ell}\,\big|\ip{\tG_{j,\ell}-G_{j,\ell},\calN_{j,\ell}(Z_{j,\ell})}_F\big|}
\le \eta\,\nu_t\,\sqrt{\Delta_t} + \eta\omega \Delta_t.
\eeqn
Similarly, using the definition of the normalization operator
$\calN_{j,\ell}$ and \req{Delta-low}, the quadratic term satisfies 
\beqn{q-term}
\frac{L_G\eta^2}{2}\sum_{\ell=1}^L\sum_{j\in\calJ_{t,\ell}}\E{\|Z_{j,\ell}\|_{*,\ell}^2\|\calN_{j,\ell}(Z_{j,\ell})\|_\ell^2}
=\frac{L_G\eta^2}{2}\sum_{\ell=1}^L\sum_{j\in\calJ_{t,\ell}}\E{\|Z_{j,\ell}\|_{*,\ell}^2}
\le \frac{L_G\eta^2}{2}\Delta_t.
\eeqn
Substituting \req{lin-term}, \req{noise-term} and \req{q-term} into
\req{descent-more} then yields that
\beqn{telesc-1}
\eta\,\E{\sum_{\ell=1}^L \tr(\Pi_{t,\ell}^{1/2})-\sum_{\ell=1}^L \tr(\Pi_{-1,\ell}^{1/2})}
\le f(X_0)-f_{\rm low} 
+ \eta\,\nu_t\,\sqrt{\Delta_t}
+ \left(\eta\omega+\frac{L_G\eta^2}{2}\right)\Delta_t,
\eeqn
which is exactly \req{master} after taking into
account that $\Pi_{-1,\ell}=\varsigma I_\ell$ for all $\ell\in\ii{L}$.
}

\noindent
We now recall from \cite{GratToin26c} a crucial consequence of the previous theorem, which
provides a bound $\Theta_t$ on the expected trace of all
preconditioners at iteration $t$.

\lthm{thm:convergence-theta}{
Suppose that  Assumptions~\ref{ass:bounded-op},
\ref{ass:smooth-op}, \ref{ass:identities} and \ref{ass:variance} hold.
Then, for all $t \ge 0$,
\beqn{Thetak-def-app}
\E{\sum_{\ell=1}^L \tr(\Pi_{t,\ell}^{1/2}) }  \le \Theta_t
\eqdef
\max\left[\,\kappa_\Theta,\,T_t\,\right],
\eeqn
where
\[
\kappa_\Theta =
\max\left[\,e^{\max\left[1,\frac{\kappa_0}{2N}\right]},\,\frac{3\kap{gap}}{\eta},\,
  24N\left(\omega+\frac{L_G}{\eta}\right)
\log\left(24N\left(\omega+\frac{L_G}{\eta}\right)\right)\right]
\]
with $\kap{gap}$ defined in Theorem~\ref{th:master-op},
\beqn{TkYk-def}
T_t =12\sqrt{N}\,
\nu_t\,\sqrt{\max\left[1,\log\left(12\sqrt{N}\,\nu_t\right)\right]},
\eeqn
$\nu_t$ being defined in Assumption~\ref{ass:variance}
and $\kappa_0 = -\sum_{\ell=1}^L d_\ell\log(d_\ell) - N\log(\varsigma)$.
}

\proof{See \cite[Lemmas 3.6, 3.7 and 3.8, and
    Theorem~3.9]{GratToin26c}. In particular, Lemma~3.6 proves that
\beqn{Delta-upper}
\Delta_t\le \kappa_0 + 2N\log\!\left(\E{\sum_{\ell=1}^L \tr(\Pi_{t,\ell}^{1/2})}\right).
\eeqn
}

\noindent
This last theorem finally gives us all the ingredients to derive the
convergence and complexity results of
Theorem~\ref{thm:dadprec-convergence}. The bound \req{Thetak-def-app},
together with Assumption~\ref{ass:opttransfer}, indeed implies
convergence of the criticality measure on each block with a {\em
  uniform rate-of-convergence bound}. As shown by the next theorem,
the expected global rate of convergence is (in order) proportional to
the value of $\Theta_t$. The proof of this statement is substantially
more involved than \cite[Theorem 3.10]{GratToin26c} because of the
introduction of delays.

\subsection*{Proof of Theorem~\ref{thm:dadprec-convergence}}

Assumption ~\ref{ass:opttransfer} gives that, for each $\ell\in\ii{L}$,
\[
\sum_{j\in\calJ_{t,\ell}} \|\tG_{j,\ell}\|_{*,\ell}^2   
\le \kappa_\circ^2\sum_{j\in\calJ_{t,\ell}}\!\tr(\calU_{j,\ell}(\tG_{j,\ell})^2)
= \kappa_\circ^2\,\tr\!\Big(\!\!\sum_{j\in\calJ_{t,\ell}}\calU_{j,\ell}(\tG_{j,\ell})^2\Big)
\le \kappa_\circ^2 \,\tr(\Pi_{t,\ell})
\]
But
\[
\tr(\Pi_{t,\ell})
=\sum_{i=1}^{d_\ell}\left(\sqrt{\lambda_i[\Pi_{t,\ell}]}\right)^2
\le \left(\sum_{i=1}^{d_\ell}\sqrt{\lambda_i[\Pi_{t,\ell}]}\right)^2
=\tr(\Pi_{t,\ell}^{1/2})^2,
\]
and thus, summing over $\ell\in\ii{L}$ and using the fact that
$\sum_i a_i^2 \le (\sum_i a_i )^2$,
\[
\sqrt{\sum_{\ell=1}^L\sum_{j\in\calJ_{t,\ell}} \|\tG_{j,\ell}\|_{*,\ell}^2}
\le \kappa_\circ \sqrt{ \sum_{\ell=1}^L\,\tr(\Pi_{t,\ell}^{1/2})^2}
\le \kappa_\circ \sqrt{\left(\sum_{\ell=1}^L\,\tr(\Pi_{t,\ell}^{1/2})\right)^2}
=  \kappa_\circ\sum_{\ell=1}^L\,\tr(\Pi_{t,\ell}^{1/2}).
\]
Using now the Cauchy-Schwarz and Jensen's
inequalities with this last bound, we deduce that
\beqn{therate}
\begin{aligned}
\sum_{\ell=1}^L\sum_{j\in\calJ_{t,\ell}} \E{\|\tG_{j,\ell}\|_{*,\ell}}
&=\E{\sum_{\ell=1}^L\sum_{j\in\calJ_{t,\ell}}\|\tG_{j,\ell}\|_{*,\ell}}\\
&\le \sqrt{L\max_{\ell\in\ii{L}}|\calJ_{t,\ell}|}\,\E{\sqrt{\sum_{\ell=1}^L\sum_{j\in\calJ_{t,\ell}}\|\tG_{j,\ell}\|_{*,\ell}^2}}\\
&\le  \kappa_\circ\sqrt{L(t+1)} \,\E{\sum_{\ell=1}^L\,\tr(\Pi_{t,\ell}^{1/2})}\\
&\le  \kappa_\circ\sqrt{L(t+1)} \,\Theta_t
\end{aligned}
\eeqn
where the last inequality results from  \req{Thetak-def-app}. 
Now let
$
H_j = ( G_{\pi(j,1),1}, \ldots, G_{\pi(j,L),L}).
$
Then
\beqn{twoterms}
\sum_{j=0}^t \E{\|G_j\|_*} \le \sum_{j=0}^t \E{\|G_j-H_j\|_*} + \sum_{j=0}^t \E{\|H_j\|_*}.
\eeqn
Consider the two terms of the right-hand side separately.
To bound the first, we apply Assumption~\ref{ass:smooth-op} and obtain that
\[
\|G_j-G_{\pi(j,\ell)}\|_*^2
\le L_G^2 \|X_j-X_{\pi(j,\ell)}\|^2
\le L_G^2\eta^2 \sum_{i=\pi(j,\ell)}^{j-1} \|Z_i\|_*^2.
\]
Assumption~\ref{ass:delays} then guarantees that each
$\|Z_i\|_*^2$ can only appear $\tau$ times when summing over $j$,
and thus that
\[
\sum_{j=0}^t\sum_{\ell=1}^L\E{\|G_j-G_{\pi(j,\ell)}\|_*^2}
\le \tau L_G^2 \eta^2 \sum_{j=0}^t\sum_{\ell=1}^L \E{\|Z_j\|_*^2}
= \tau L_G^2 \eta^2 L \sum_{j=0}^t \E{\|Z_j\|_*^2}.
\]
But \req{Delta-low}, \req{Delta-upper} and \req{Thetak-def-app} imply that
\beqn{bdDelta}
\sum_{j=0}^t\sum_{\ell=1}^L \E{\|Z_{j,\ell}\|_{*,\ell}^2}
\le \Delta_t
\le \kappa_0+2N\log\!\left(\E{\sum_{\ell=1}^L\tr(\Pi_{t,\ell}^{1/2})}\right)
\le \kappa_0+2N\log(\Theta_t).
\eeqn
and thus, using \req{big-dual-norm}, that
\[
\sum_{j=0}^t\E{\|G_j-H_j\|_*^2}
=\sum_{j=0}^t\sum_{\ell=1}^L\E{\|G_{j,\ell}-G_{\pi(j,\ell),\ell}\|_{*,\ell}^2}
\le \tau L_G^2 \eta^2 L \left[\kappa_0+2N\log(\Theta_t)\right].
\]
The Cauchy-Schwarz inequality then gives that
\beqn{bdt1}
\sum_{j=0}^t\E{\|G_j-H_j\|_*}
\le \sqrt{t+1}\sqrt{\sum_{j=0}^t\E{\|G_j-H_j\|_*^2}}
\le  \sqrt{t+1}\,\sqrt{\tau L_G^2\eta^2 L \left[\kappa_0+2N\log(\Theta_t)\right]}.
\eeqn
Consider now the second term in the right-hand side of \req{twoterms}. We
have, again using \req{big-dual-norm}, that
\[
\sum_{j=0}^t \|H_j\|_*
= \sum_{j=0}^t \sqrt{\sum_{\ell=1}^L\|G_{\pi(j,\ell),\ell}\|_{*,\ell}^2}
\le  \sum_{j=0}^t\sum_{\ell=1}^L\|G_{\pi(j,\ell),\ell}\|_{*,\ell}
\]
Now Assumption~\ref{ass:delays} ensures each
$\|G_{\pi(j,\ell),\ell}\|_{*,\ell}^2$ appears at most $\tau$
times in the sum over $j$.  Thus,
\beqn{aa1}
\sum_{j=0}^t \|H_j\|_*
\leq \sum_{j=0}^t\sum_{\ell=1}^L\|G_{\pi(j,\ell),\ell}\|_{*,\ell}
\le \tau \sum_{\ell=1}^L\sum_{j\in\calJ_{t,\ell}}\|G_{j,\ell}\|_{*,\ell}.
\eeqn
But Assumption~\ref{ass:unbiased}, Jensen's inequality and the law of
total expectation ensure that
\[
\E{\|G_{j,\ell} \|_{*,\ell}}
= \E{\|\Econd{j}{\tG_{j,\ell}}\|_{*,\ell}}
\le \E{\Econd{j}{\|\tG_{j,\ell}\|_{*,\ell}}}
= \E{\|\tG_{j,\ell} \|_{*,\ell}}.
\]
Substituting this identity in \req{aa1} and using \req{therate}, we
then obtain that
\beqn{bdt2}
\sum_{j=0}^t \E{\|H_j\|_*}
\le  \tau
\sum_{\ell=1}^L\sum_{j\in\calJ_{t,\ell}}\E{\|\tG_{j,\ell}\|_{*,\ell}}
\le \tau \kappa_\circ\sqrt{L(t+1)} \,\Theta_t.
\eeqn
Combining \req{bdt1} and \req{bdt2} and dividing by $t+1$ finally yields \req{thetruerate}.

\subsection*{Proof of Corollary~\ref{therate2}}

Observe that, when $\Theta_t$ (as defined in
Theorem~\ref{thm:convergence-theta}) grows with $t$, the bound 
\req{thetruerate} is dominated by the term $\kappa_c\Theta_t/\sqrt{t+1}$.
Since Corollary~\ref{therate2} solely aims at describing dominating terms
for growing $t$, we may ignore the other terms and merely consider
that \req{thetruerate} reduces to
\beqn{simple-conv}
 \frac{1}{t+1}\sum_{j=0}^t\E{\|G_j\|_*}
\le \frac{\kappa_c \Theta_t}{\sqrt{t+1}},
\eeqn
which is identical to the result (equation (3.42)) of Theorem~3.10 in \cite{GratToin26c}.
The rest of the proof then follows that of
\cite[Corollary~3.11]{GratToin26c} with $\nu_k^2$ replaced by
\[
\mu_{\max}\max_{\ell\in\ii{L}}\left[\sum_{j\in\calJ_{t,\ell}}\frac{1}{\min[|\calJ_{j,\ell}|]}\right]
\le \left\{\begin{array}{ll}
            \bigfrac{\mu_{\max} \sigma_{\rm tot}^2}{1-\alpha}\psi_t^{1-\alpha} &
            \tim{if } \alpha < 1,\\
            \mu_{\max} \sigma_{\rm tot}^2 \log(\psi_t) & 
            \tim{if } \alpha = 1,\\
            \mu_{\max} \sigma_{\rm tot}^2\zeta(\alpha) &
            \tim{if } \alpha > 1,
            \end{array}\right.
\]
yielding \req{therate2-psi}.

\appnumsection{Appendix~2: Detailed analysis for \lasname\ with momentum}

\noindent
We start by establishing a bound on the norm of the
difference between the block-wise momentum $M_{t,\ell}$ and the
approximate gradient $\tG_{t,\ell}$. 

\llem{lem:E-form}{Consider Algorithms~\asnameMM\ and \asnameMG.
Suppose that Assumption~\ref{ass:smooth-op} holds. 
Define
\vspace*{-1mm}
\beqn{E-def}
E_{t,\ell} = M_{t,\ell}-\tG_{t,\ell} \tim{ for } t\ge 0 \tim{and} \ell\in\calA_t.
\vspace*{-2mm}
\eeqn
\vspace*{-2mm}
Then
\beqn{errEsq}
\sum_{\ell=1}^L\sum_{j\in\calJ_{t,\ell}}\E{\|E_{j,\ell}\|_{*,\ell}^2}
\!\le \!\frac{3}{(1\!-\!\mu_{\max})^2}\!\!\left[2
  \sum_{\ell=1}^L\sum_{j\in\calJ_{t,\ell}}\bar{\mu}_{j,\ell}^2\E{\|\tG_{j,\ell}-G_{j,\ell}\|_{*,\ell}^2}
  \!+\!\tau L L_G^2\eta^2\sum_{j=0}^t\E{\|Z_j\|_*^2}\right]
\eeqn
where $\bar{\mu}_{j,\ell} = \max[\,\mu_{\pi(j,\ell),\ell},\,\mu_{j,\ell}\,]$.
}

\proof{
Suppose that $t\ge1$  and note that the definition
of the momentum and \req{E-def} give that, for $\ell\in\calA_t$,
\[
\begin{aligned}
E_{t,\ell}
& = \mu_{t,\ell} M_{\pi(t-1,\ell),\ell} +(1-\mu_{t,\ell}) \tG_{t,\ell}-\tG_{t,\ell}\\
& = \mu_{t,\ell}(M_{\pi(t-1,\ell),\ell}-\tG_{\pi(t-1,\ell),\ell})
    +\mu_{t,\ell}(\tG_{\pi(t-1,\ell),\ell}-G_{\pi(t-1,\ell),\ell}) + \mu_{t,\ell}(G_{\pi(t-1,\ell),\ell}-\tG_{t,\ell})\\
& = \mu_{t,\ell}E_{\pi(t-1,\ell),\ell} +\mu_{t,\ell}(\tG_{\pi(t-1,\ell),\ell}-G_{\pi(t-1,\ell),\ell}) + \mu_{t,\ell}(G_{\pi(t-1,\ell),\ell}-\tG_{t,\ell})\\
\end{aligned}
\]
Thus, using
the facts that $\mu_{t,\ell} \le \mu_{\max}< 1$, that
$(a+b)^2\le (1+\rho)a^2+ (1+1/\rho)b^2$ for any $\rho>0$ and that
$(a+b+c)^2 \le 3a^2 +3b^2+3c^2$, we obtain that
\[
\begin{aligned}
\|E_{t,\ell}\|_{*,\ell}^2
\le&  (1+\rho)\mu_{\max}^2\|E_{\pi(t-1,\ell),\ell}\|_{*,\ell}^2 \\
&+ 3\left(1+\frac{1}{\rho}\right)\mu_{t,\ell}^2
\left[\|G_{t,\ell}-G_{\pi(t-1,\ell),\ell}\|_{*,\ell}^2
  +\|\tG_{\pi(t-1,\ell),\ell}-G_{\pi(t-1,\ell),\ell}\|_{*,\ell}^2+\|\tG_{t,\ell}-G_{t,\ell}\|_{*,\ell}^2\right].
\end{aligned}
\]
Now let $\rho = (1-\mu_{\max})/\mu_{\max}$. Then $(1+\rho)\mu_{\max}^2 = \mu_{\max} < 1$ and
$(1+1/\rho)=1/(1-\mu_{\max})$. Hence, summing over $j\in\calJ_{t,\ell}$, we obtain that
\[
\begin{aligned}
\sum_{\mystack{j\in\calJ_{t,\ell}}{j\ge1}}\|E_{j,\ell}\|_{*,\ell}^2
\le & \mu_{\max}\sum_{j\in\calJ_{k,\ell}}\|E_{\pi(j-1,\ell),\ell}\|_{*,\ell}^2\\*[-2ex]
& \hspace*{-15mm}+ \frac{3}{1-\mu_{\max}}\sum_{j\in\calJ_{t,\ell}}\mu_{j,\ell}^2
\left[\|G_{j,\ell}-G_{\pi(j-1,\ell),\ell}\|_{*,\ell}^2
  +\|\tG_{\pi(j-1,\ell),\ell}-G_{\pi(j-1,\ell),\ell}\|_{*,\ell}^2+\|\tG_{j,\ell}-G_{j,\ell}\|_{*,\ell}^2\right]\\
\end{aligned}
\]
Observe now that \req{big-dual-norm}, Assumption~\ref{ass:smooth-op} with \req{xkp1} and
the definition of the normalization operator imply that
\beqn{isLip}
\begin{aligned}
\|G_{j,\ell}-G_{\pi(j-1,\ell),\ell}\|_{*,\ell}^2
&\le \|G_j-G_{\pi(j-1,\ell)}\|_*^2\\
&\le L_G^2\sum_{q=1}^L\|X_{j,q}-X_{\pi(j-1,\ell),q}\|_q^2\\
&\le L_G^2\sum_{q=1}^L\sum_{s=\pi(j-1,\ell)}^{j-1}\|X_{s+1,q}-X_{s,q}\|_q^2\\
&=L_G^2\eta^2\sum_{q=1}^L\sum_{s=\pi(j-1,\ell)}^{j-1}\|Z_{s,q}\|_{*,q}^2\|\calN_{s,q}(Z_{s,q})\|_q^2\\
&=L_G^2\eta^2\sum_{q=1}^L\sum_{s=\pi(j-1,\ell)}^{j-1}\|Z_{s,q}\|_{*,q}^2.
\end{aligned}
\eeqn
But Assumption~\ref{ass:delays} implies that the sum over $s$ in the last right-hand side
contains at most $\tau$ terms. Thus, using also \req{big-dual-norm}
and the fact that $\mu_{j,\ell}<1$,
\[
\sum_{\mystack{j\in\calJ_{t,\ell}}{j\ge 1}}\mu_{j,\ell}^2\|G_{j,\ell}-G_{\pi(j-1,\ell),\ell}\|_{*,\ell}^2
\le \tau L_G^2\eta^2 \sum_{q=1}^L\sum_{s=0}^{t-1}\|Z_{s,q}\|_{*,q}^2
= \tau L_G^2\eta^2\sum_{s=0}^{t-1}\|Z_s\|_*^2.
\]
Summing now over $\ell\in\ii{L}$, taking full
expectation and defining
$\vartheta_{t,\ell}^2=\E{\|\tG_{t,\ell}-G_{t,\ell}\|_{*,\ell}^2}$,
we deduce that
\[
\begin{aligned}
\sum_{\ell=1}^L&\sum_{\mystack{j\in\calJ_{t,\ell}}{j\ge1}}\E{\|E_{j,\ell}\|_{*,\ell}^2}
\le\mu_{\max}\sum_{\ell=1}^L\sum_{\mystack{j\in\calJ_{t,\ell}}{j\ge1}}\E{\|E_{\pi(j-1,\ell),\ell}\|_{*,\ell}^2}\\
&\hspace*{23mm} +\frac{3}{1-\mu_{\max}}\Bigg[\tau L_G^2\eta^2\sum_{\ell=1}^L\sum_{s=0}^{t-1}\E{\|Z_s\|_*^2}
     + \sum_{\ell=1}^L\sum_{\mystack{j\in\calJ_{t,\ell}}{j\ge1}}\mu_{j,\ell}^2(\vartheta_{\pi(j-1,\ell),\ell}^2+\vartheta_{j,\ell}^2)\Bigg]\\
&\le\mu_{\max}\sum_{\ell=1}^L\sum_{\mystack{j\in\calJ_{t,\ell}}{j\ge1}}\E{\|E_{j,\ell}\|_{*,\ell}^2}
    +\frac{3}{1-\mu_{\max}}\Bigg[\tau L L_G^2\eta^2\sum_{s=0}^{t-1}\E{\|Z_s\|_*^2}
    + 2\sum_{\ell=1}^L\sum_{\mystack{j\in\calJ_{t,\ell}}{j\ge1}}\bar{\mu}_{j,\ell}^2\vartheta_{j,\ell}^2 \Bigg],
\end{aligned}
\]
so that
\[
\begin{aligned}
\sum_{\ell=1}^L\sum_{\mystack{j\in\calJ_{t,\ell}}{j\ge1}}\E{\|E_{j,\ell}\|_{*,\ell}^2}
&\le \frac{3}{(1-\mu_{\max})^2}\Bigg[\tau L L_G^2\eta^2\sum_{s=0}^{t-1}\E{\|Z_s\|_*^2}
  + 2\sum_{\ell=1}^L\sum_{\mystack{j\in\calJ_{t,\ell}}{j\ge1}}\bar{\mu}_{j,\ell}^2\vartheta_{j,\ell}^2\Bigg]\\
\end{aligned}
\]
which, with the fact that $E_0= \tG_0-G_0$ and \req{big-dual-norm}, concludes
the proof of \req{errEsq}.
}

\subsection*{Proof of Theorem~\ref{thm:momMM-convergence}}

Observe now that the \asnameMM\ algorithm
is nothing but Algorithm \asname\ where the momentum $M_{t,\ell}$
plays the role of the approximate gradient $\tG_{t,\ell}$.  Moreover
Assumption~\ref{ass:mom-identities} exactly reflects this change of
perspective. Also note that, for all $\ell\in\ii{L}$ and $j\in\calJ_{t,\ell}$,
\[
\|M_{j,\ell} - G_{j,l}\|_{*,\ell}^2
\le 2 \|E_{j,\ell}\|_{*,\ell}^2 +  2 \|\tG_{j,\ell} - G_{j,l}\|_{*,\ell}^2.
\]
Hence, using Lemma~\ref{lem:E-form} and  given that
$\bar{\mu}_{j,\ell}<1$, we have that
\[
\sum_{\ell=1}^L\sum_{j\in\calJ_{t,\ell}}\E{\|M_{j,\ell}-G_{j,l}\|_{*,\ell}^2}
<\left(\frac{3}{(1\!-\!\mu_{\max})^2}+2\right)\sum_{j=0}^t\E{\|\tG_j-G_j\|_*^2}
+\frac{2[\tau L  L_G^2\eta^2}{(1\!-\!\mu_{\max})^2}\sum_{j=0}^t\E{\|Z_j\|_*^2}.
\]
This provides an adapted formulation to replace \req{var-cond} in
Assumption~\ref{ass:variance}, which thus continues to hold with
\beqn{nuknown}
\nu_t^2 = \left(\frac{3}{(1-\mu_{\max})^2}+2\right)\sum_{j=0}^t\E{\|\tG_j-G_j\|_*^2}
\tim{ and }
\omega^2 = \frac{2\tau L L_G^2\eta^2}{(1-\mu_{\max})^2}.
\eeqn
As a consequence, the theory developed for the \asname\ algorithm also holds for
its \asnameMM\ variant and this proves
Theorem~\ref{thm:momMM-convergence}.

\subsection*{Proof of Theorem~\ref{thm:convergence-alt}}

This section considers the convergence of the \asnameMG\ algorithm.
We first recall a result on the impact of using this algorithm on the
structural identities of Assumption~\ref{ass:mom-identities}.

\llem{lem:perturbs}{
Suppose that Assumptions~\ref{ass:smooth-op} and \ref{ass:mom-identities}
and \req{ass:sub-add} hold.
Then
\beqn{pineq1}
\!\|Z_{t,\ell}\|_{*,\ell}\,\ip{G_{t,\ell},\calN_{t,\ell}(Z_{t,\ell})}_F
\ge \!\frac{1}{\kappa_\Box}\tr\!\big(\Pi_{t,\ell}^{-1/2}\calU_{t,\ell}(\tG_{t,\ell})^2\big)
-\tr\!\big(\Pi_{t,\ell}^{-1/2}\calU_{t,\ell}(E_{t,\ell})^2\big)
-\|Z_{t,\ell}\|_{*,\ell}\|E_{t,\ell}\|_{*,\ell}
\eeqn
and
\beqn{pineq2}
\|Z_{t,\ell}\|_{*,\ell}^2
\le \kappa_\Box \!\tr\!\big(\Pi_{t,\ell}^{-1}\calU_{t,\ell}(G_{t,\ell})^2\big)
    + \kappa_\Box \tr\!\big(\Pi_{t,\ell}^{-1}\calU_{t,\ell}(E_{t,\ell})^2\big)
\eeqn
}
\proof{See \cite[Lemmas A.1 and A.2]{GratToin26c}.}

\llem{lem:pert-bounds}{
  Suppose that that Assumption~\ref{ass:smooth-op} and \ref{ass:mom-identities}
  and \req{ass:sub-add} hold.  Then
  \beqn{prod-bound}
  \sum_{\ell=1}^L\sum_{j\in\calJ_{t,\ell}}\E{\|Z_{j,\ell}\|_{*,\ell}\|E_{j,\ell}\|_{*\ell}}
  \le \frac{12\theta_t^2}{(1-\mu_{\max})^2}
    +\left(\frac{6\tau L L_G^2\eta^2}{(1-\mu_{\max})^2}+2\right)\sum_{j=0}^t\E{\|Z_j\|_*^2},   
  \eeqn
  \beqn{pert1-bound}
  \begin{aligned}
  \sum_{j=0}^t\sum_{\ell=1}^L\E{\tr\!\big(\Pi_{j,\ell}^{-1/2}\calU_{t,\ell}(E_{j,\ell})^2\big)}
  &\le \frac{6\kappa_\diamond\theta_t^2}{(1-\mu_{\max})^2\sqrt{\varsigma}}
     + \frac{3\kappa_\diamond \tau L L_G^2\eta^2}{(1-\mu_{\max})^2\sqrt{\varsigma}}\,
       \sum_{j=0}^t\E{\|Z_j\|_*^2}
   \end{aligned}
  \eeqn
  \vspace*{-2mm}
  and
  \beqn{pert2-bound}
  \sum_{j=0}^t\sum_{\ell=1}^L\E{\tr\!\big(\Pi_{t,\ell}^{-1}\calU_{t,\ell}(E_{t,\ell})^2\big)}
  \le \frac{6\kappa_\diamond\theta_t^2}{(1-\mu_{\max})^2\varsigma}
       + \frac{3\kappa_\diamond \tau L L_G^2\eta^2}{(1-\mu_{\max})^2\varsigma}\,
       \sum_{j=0}^t\E{\|Z_j\|_*^2}.
  \eeqn
}

\proof{See \cite[Lemma A.3]{GratToin26c}.}

\llem{lem:pert-bounds2}{
Suppose that Assumption~\ref{ass:smooth-op} and \ref{ass:mom-identities} and
\req{ass:sub-add} hold. Suppose in addition that \req{small-eta} is satisfied.
Then
\beqn{bound-p1}
\begin{aligned}
\bigsum_{j=0}^t\bigsum_{\ell=1}^L\E{\|Z_{j,\ell}\|_{*,\ell}\,\ip{G_{j,\ell},\calN_{j,\ell}(Z_{j,\ell})}_F}
\ge &\bigfrac{1}{\kappa_\Box}\bigsum_{j=0}^t\bigsum_{\ell=1}^L\E{\tr\!\big(\Pi_{t,\ell}^{-1/2}\calU_{t,\ell}(\tG_{t,\ell})^2\big)}\\
&-\kappa_{1,\nu}\,\theta_t^2-
\kappa_{1,z}\,\bigsum_{j=0}^t\E{\|Z_j\|_*^2}
\end{aligned}
\eeqn
\vspace*{-3mm}%
with
\beqn{kappa1}
\kappa_{1,\nu} = \frac{6\kappa_\diamond}{(1-\mu_{\max})^2\sqrt{\varsigma}}
+ \frac{12}{(1-\mu_{\max})^2}
\tim{and}
\kappa_{1,z} =
\frac{3\kappa_\diamond \tau L L_G^2\eta^2}{(1-\mu_{\max})^2\sqrt{\varsigma}}
+\frac{6\tau L L_G^2\eta^2}{(1-\mu_{\max})^2}+2,
\eeqn
and
\beqn{bound-p2}
\sum_{j=0}^t\E{\|Z_j\|_*^2}
\le 2 \sum_{j=0}^t\sum_{\ell=1}^L\E{\tr\!\big(\Pi_{j,\ell}^{-1}\calU_{j,\ell}(G_{j,\ell})^2\big)}
    + \kappa_{2,\nu}\,\theta_t^2+ \kappa_{2,z},
\eeqn
\vspace*{-3mm}%
where
\beqn{kappa2}
\kappa_{2,\nu} =
\frac{6\kappa_\Box\kappa_\diamond}{(1-\mu_{\max})^2\varsigma}
\tim{ and }
\kappa_{2,z} = \left\{\begin{array}{ll}
0
&\tim{if the first part of \req{small-eta} holds,}\\
\frac{3\kappa_\Box\kappa_\diamond
  \tau L L_G^2\eta^2}{(1-\mu_{\max})^2\varsigma}\,\kappa_Z 
&\tim{if the second part of \req{small-eta} holds.}
\end{array}\right.
\eeqn
}

\proof{
The inequality \req{bound-p1} is obtained by substituting
\req{prod-bound} and \req{pert1-bound} into \req{pineq1},
Moreover, taking the expectation in \req{pineq2}, summing for
$j\in\iiz{t}$ and $\ell\in\ii{L}$ and substituting \req{pert2-bound} gives that
\[
\begin{aligned}
\sum_{j=0}^t\sum_{\ell=1}^L \E{\|Z_{j,\ell}\|_{*,\ell}^2}
&\le \sum_{j=0}^t\sum_{\ell=1}^L \E{\tr\!\big(\Pi_{j,\ell}^{-1}\calU_{j,\ell}(G_{j,\ell})^2\big)}
    +
    \frac{6\kappa_\Box\kappa_\diamond}{(1-\mu_{\max})^2\varsigma}\,\theta_t^2\\
    & \hspace*{2cm}
    + \frac{3\kappa_\Box\kappa_\diamond \tau L L_G^2\eta^2}{(1-\mu_{\max})^2\varsigma}\,
    \sum_{j=0}^t\E{\|Z_j\|_*^2}.
\end{aligned}
\]
Suppose first that the first part of \req{small-eta} holds. Then
\[
\frac{3\kappa_\Box\kappa_\diamond \tau L L_G^2\eta^2}{(1-\mu_{\max})^2\varsigma} \le \frac{1}{2}
\]
and \req{bound-p2} follows with $\kappa_{2,z}=0$. Alternatively, if
the second part of \req{small-eta} holds, then \req{bound-p2} follows
with
\[
\kappa_{2,z}=\frac{3\kappa_\Box\kappa_\diamond \tau L
  L_G^2\eta^2}{(1-\mu_{\max})^2\varsigma}\,\kappa_{\mu Z}.
\]
}

\noindent
From this point on, the analysis follows the lines of the argument of
Appendix~1 with
Lemma~\ref{lem:pert-bounds2} providing an alternate set of 
structural inequalities. Lemma~\ref{lem:sqrt-op}
is unchanged. The modifications to Theorem~\ref{th:master-op} are
minor.  It is restated as follows.

\lthm{th:master-op-mom}{
Suppose that Assumption~\ref{ass:bounded-op} and
\ref{ass:mom-identities} and \req{small-eta} hold.
Define $\Delta_j$ as in \req{Delta-def}.
Then, for every $k\ge0$,
\beqn{master-mom}
\eta\,\E{\sum_{\ell=1}^L \tr(\Pi_{t,\ell}^{1/2})}
\le \kap{gap}+\kappa_{\nu\nu}\theta_t^2+\eta\,\kappa_{\nu\Delta}\,\theta_t\,\sqrt{\Delta_t}+\eta\,\kappa_\Delta\,\Delta_t,
\eeqn
where
\beqn{kaps-def1}
\kap{gap} \eqdef \E{f(X_0)}-f_{\rm low} +
\eta\,\varsigma\,N + \eta\sqrt{\kappa_{2,z}}+\left(\eta\kappa_{1,z}+\frac{L_G\eta^2}{2}\right)\,\kappa_{2,z}
\eeqn
\beqn{kaps-def2}
\kappa_{\nu\nu} =
\eta\left(\kappa_{1,\nu}+\sqrt{\kappa_{2,\nu}}+\kappa_{1,z}\kappa_{2,\nu}+\frac{L_G\eta}{2}\right),
\ms
\kappa_{\nu\Delta} = \sqrt{2}
\tim{ and }
\kappa_\Delta = 2\kappa_{1,z}+L_G\eta.
\eeqn
}

\proof{
In the proof of Theorem~\ref{th:master-op}, a perturbation
$\eta\kappa_{1,\nu}\,\theta_t^2
+\eta\kappa_{1,z}\,\sum_{j=0}^t\sum_{\ell=1}^L\E{\|Z_{t,\ell}\|_{*,\ell}^2}$
is subtracted from the right-hand side of
\req{lin-term} in order to reflect \req{bound-p1}. The proof then
goes on unmodified, up to the substitution leading to \req{telesc-1},
in which the perturbed \req{lin-term} then gives that
\[
\eta\,\E{\sum_{\ell=1}^L\tr(\Pi_{t,\ell}^{1/2})-\sum_{\ell=1}^L\tr(\Pi_{-1,\ell}^{1/2})}
\le f(X_0)-f_{\rm low}
+\eta\kappa_{1,\nu} \theta_t^2
+\eta\theta_t\sqrt{\zeta_t}+\left(\eta\kappa_{1,z}+\frac{L_G\eta^2}{2}\right)\,\zeta_t.
\]
where $\zeta_t = \sum_{j=0}^t\sum_{\ell=1}^L\E{\|Z_{j,\ell}\|_{*,\ell}^2}$.
But this definition, \req{bound-p2} and \req{log-pot} give that
\[
\zeta_t
\le \kappa_{2,z}+\kappa_{2,\nu}\,\theta_t^2 
+ 2\sum_{j=0}^t\sum_{\ell=1}^L\E{\tr\!\big(\Pi_{j,\ell}^{-1}\calU_{j,\ell}(G_{j,\ell})^2\big)}
\le \kappa_{2,z}+\kappa_{2,\nu}\,\theta_t^2 + 2\Delta_t,
\]
and thus, using $\sqrt{a+b}\le \sqrt{a}+\sqrt{b}$,
\[
\begin{aligned}
\eta\,\E{\sum_{\ell=1}^L\tr(\Pi_{t,\ell}^{1/2})-\sum_{\ell=1}^L\tr(\Pi_{-1,\ell}^{1/2})}
& = f(X_0)-f_{\rm low} +\eta\sqrt{\kappa_{2,z}}
+\eta\left(\kappa_{1,\nu}+\sqrt{\kappa_{2,\nu}}+\kappa_{1,z}\kappa_{2,\nu}+\frac{L_G\eta}{2}\right)\theta_t^2\\
&+  \sqrt{2}\eta\theta_t\sqrt{\Delta_t}
+2\left(\eta\kappa_{1,z}+\frac{L_G\eta^2}{2}\right)\,\Delta_t + \left(\eta\kappa_{1,z}+\frac{L_G\eta^2}{2}\right)\,\kappa_{2,z}
\end{aligned}
\]
yielding \req{master-mom}-\req{kaps-def2} after taking into
account that $\Pi_{-1,\ell}=\varsigma I_\ell$ for all
$\ell\in\ii{L}$.
}

\noindent
Note that the form of bound \req{master-mom} differs very little from that
of \req{master}: besides using different constants, \req{master-mom}
now involves a term in $\kappa_{\nu\nu}\theta_t^2$. This term
then percolates through the proof of
Theorem~\ref{thm:convergence-theta}, so that \req{Thetak-def-app} 
now becomes
\beqn{Thetak-def-app-MG}
\E{\sum_{\ell=1}^L \tr(\Pi_{t,\ell}^{1/2}) }  \le \Theta_t
\eqdef
\max\left[\,\kappa_\Theta,\,\frac{3\kappa_{\nu\nu}\theta_t^2}{\eta},\,T_t\,\right].
\eeqn
The proof of Theorem~\ref{thm:dadprec-convergence} then stands without
modification other than using this updated value of $\Theta_t$,
finally yielding Theorem~\ref{thm:convergence-alt}.

\subsection*{Proof of Corollary~\ref{therate2-alt}}

Finally, the proof of Corollary~\ref{therate2-alt} is similar in
spirit to that of Corollary~3.11 in \cite{GratToin26c}, with two
twists.  The first is that the formula for the block-wise
variance is now given by \req{var-cond-alt2}, which, together with
\req{mutell}, implies\footnote{With the convention that
  $|\calJ_{-1,\ell}| = 1$ for all $\ell\in\ii{L}$.} that
\[
\begin{aligned}
\sum_{\ell=1}^L\sum_{j\in\calJ_{t,\ell}}\|M_{j,\ell}-G_{j,\ell}\|_{*,\ell}^2
& \le \mu_{\max}\,\sum_{\ell=1}^L  \sigma_\ell^2\,
\sum_{j\in\calJ_{t,\ell}}\frac{1}{\min[|\calJ_{j,\ell}|,|\calJ_{j-1,\ell}|]^{\alpha+2\beta}}
+\omega^2\sum_{j=0}^t\E{\|Z_{j,\ell}\|_{*,\ell}^2}\\
& \le \mu_{\max} \sigma_{\rm tot}^2\,
\sum_{j=1}^t\frac{1}{|\calJ_{j-1,\ell}|^{\alpha+2\beta}}+\omega^2\sum_{j=0}^t\E{\|Z_{j,\ell}\|_{*,\ell}^2}.
\end{aligned}
\]
Therefore, using $\psi_{t-1}\le\psi_t$, we may set the bound
\beqn{thetaMG}
\theta_t^2= \left\{\begin{array}{ll}
            \frac{\sigma_{\rm tot}^2}{1-\alpha-2\beta}\psi_t^{1-\alpha-2\beta} &
            \tim{if } \alpha+2\beta < 1,\\
            \sigma_{\rm tot}^2 \log(\psi_t) & 
            \tim{if } \alpha+2\beta = 1,\\
           \sigma_{\rm tot}^2\zeta(\alpha+2\beta) &
            \tim{if } \alpha+2\beta > 1.\\
            \end{array}\right.
\eeqn
The second is the presence of term $3\kappa_{\nu\nu}\theta_t^2$
in \req{Thetak-def-alt}.
Since the dominant term in the convergence bound \req{thetruerate-alt}
is the last, the final order of convergence is determined by the
maximal power of $\theta_t$ occuring in the expression of $\Theta_t$,
which is now $\theta_t^2$ because of this new term. Thus substituting \req{thetaMG} in
\req{thetruerate-alt} and only keeping the dominant term in $t$ gives
\req{therateMG}.
\epr

\appnumsection{Appendix~3: Implementation and additional results for \\asynchronous momentum variants}

We first describe our warm-started power iteration to compute an
approximation of the nuclear norm of $G$.  It is based on the proxy
\[
\|G\|_* \approx \frac{\|G\|_F^2}{\sigma_{\max}[G]}
\]
where $\sigma_{\max}[G]$ is the largest singular value of $G$.  Note
that $\|G\|_F^2$ is easily computable, but we estimate
$\sigma_{\max}[G]$ using two iterations of the power iterations on
$G$. This does not ensure a convergent estimator, but gives a
deterministic lower bound
\[
\frac{\|G\|_F^2}{\sigma_{\max}[G]}
= \frac{\sum_{i=1}^n \sigma_i[G]^2}{\sigma_{\max}[G]}
\le \sum_{i=1}^n \sigma_i[G] = \|G\|_*,
\]
where equality holds if and only if all singular values are identical.
Our algorithm is warm-started: we initialize the
power iteration with the result obtained at the previous optimisation
iteration. As the gradient $G$ changes progressively, so does its
dominant singular vector, and two iterations are typically enough to
obtain an accurate estimation est$(\|G\|_*)$.
The detailed algorithm is given \vpageref{algo:power}.

\algo{algo:power}{\al{WSPOWER}}{
  Given: a starting vector $v$ resulting from the previous
  optimization iteration (a random vector at the first).
  \begin{enumerate}
  \item Compute $u = Gv$, $u= u/\|u\|$, $v = G^Tu$, $\sigma_{\max} = \|v\|$,
    $v=v/\|v\|$ \hspace*{1cm}(repeat twice)
    \vspace*{-2.5mm}
  \item Set est$(\|G\|_*) =
    \bigfrac{1}{\sigma_{\max}}\bigsum_{i=1}^m\bigsum_{j=1}^m G_{i,j}^2$,
    \vspace*{-2.5mm}
  \item Save $v$ for the next call.
  \end{enumerate}
}

The cost of using \al{WSPOWER} for an $m \times n$ matrix amounts to
$\calO(4mn)$ flops for the four matrix-vector products,
plus $\calO(mn)$ flops for $\|G\|_F^2$, giving a total of
$\calO(5mn)$ flops, compared
to $\calO(mn\min(m,n))$ flops for the SVD. In our experiments with $256\times
2048$ matrices, using \al{WSPOWER} is $\min(m,n)/5 \approx 50$ times faster.

Although a mere proxy of $\|G\|_*$, \al{WSPOWER} works remarkably
well, but its warm-start feature is crucial. Without it, accuracy results diverge on larger models
(SVHN: 56\%, MoE-FMNIST: 77\%, to compare with SVHN 70.44\% ± 0.43,
MoE 88.63\% ± 0.34 with warm start, this last result being the best
achieved for this benchmark in all our runs)
The throughput of our asynchronous {\sf MG} optimization variants using \al{WSPOWER}
then become comparable that of {\sf MM} variants, as shown
in Table~\ref{tab:mom-vars-tput} below.

We now complete the results mentioned in Section~\ref{sec:momentum}.
Table~\ref{tab:mom-vars-loss} gives the final loss values obtained
when using different asynchronous momentum variants.

\begin{table}[htbp]
\centering \footnotesize
\begin{tabular}{|l|c|c|c|c|c|c|c|}
\hline
Dataset & No Mom. & {\sf MMconst} & {\sf MGconst} & {\sf MGconstA} &{\sf MMdecay} & {\sf MGdecay} & {\sf MGdecayA}\\
\hline
Fashion-MNIST  & 0.308 & \textbf{0.239} & 0.336 & 0.259 & 0.267 & 0.316 & 0.273 \\
MoE-FMNIST     & 0.279 & 0.240 & 0.245 & \textbf{0.186} & 0.290 & 0.296 & 0.288 \\
CovType        & 0.240 & \textbf{0.149} & 0.226 & 0.173 & 0.225 & 0.256 & 0.229 \\
MovieLens      & 0.353 & \textbf{0.146} & 0.347 & 0.209 & 0.344 & 0.397 & 0.305 \\
Criteo         & 0.088 & \textbf{0.008} & 0.107 & 0.027 & 0.081 & 0.126 & 0.041 \\
SVHN           & 0.819 & 0.856 & 0.954 & 0.798 & \textbf{0.762} & 0.932 & 0.809 \\
\hline
\end{tabular}
\caption{Final loss value for asynchronous momentum variants}
\label{tab:mom-vars-loss}
\end{table}

\noindent
We see that the {\sf MMconst} variant dominates on  CovType (0.149),
Criteo (0.008!), Fashion-MNIST (0.239)and  MovieLens (0.146), while
{\sf  MGconstA} is best for MoE-FMNIST and {\sf MGdecay} for SVHN.

The generalization metrics for the different variants with momentum
are given in Table~\ref{tab:mom-vars-gen}.

\begin{table}[htbp]
\centering \footnotesize
\begin{tabular}{|l|c|c|c|c|c|c|c|c|}
\hline
Dataset & Metric& No Mom. & {\sf MMconst} & {\sf MGconst} & {\sf MGconstA} &{\sf MMdecay} & {\sf MGdecay} & {\sf MGdecayA}\\
\hline
Fashion-MNIST  & acc.(\%) & 87.34 & 88.16 & 86.31 & 87.86 & \textbf{88.19} & 87.00 & 87.92 \\
MoE-FMNIST     & acc.(\%) & 87.39 & 87.97 & 88.11 & \textbf{88.63} & 86.93 & 86.99 & 87.11 \\
CovType        & acc.(\%) & 89.56 & \textbf{92.71} & 90.01 & 91.92 & 90.13 & 88.88 & 90.07 \\
MovieLens      & RMSE     & 0.916 & 0.992 & \textbf{0.914} & 0.956 & 0.915 & 0.923 & 0.926 \\
Criteo         & AUC      & 0.717 & 0.720 & \textbf{0.728} & 0.716 & 0.716 & 0.723 & 0.708 \\
SVHN           & acc.(\%) & 69.39 & 68.69 & 65.10 & 70.44 & \textbf{71.55} & 66.19 & 69.45 \\
\hline
\end{tabular}
\caption{Generalization metrics for asynchronous momentum variants}
\label{tab:mom-vars-gen}
\end{table}

Care should however be exercised: compared to the momentum-less
option, the loss for MovieLens improves from 0.353 to 0.008, but the
accuracy worsens from 0.916 to 0.992, illustrating a possible overfitting
in that case.  

The throughputs per problem are given in
Table~\ref{tab:mom-vars-tput}, which shows that the introduction of
the {\sf MM} momentum has a limited effect on throughput, but that the
{\sf MG} approach has a clear negative impact, due to the double
decomposition that we pinpointed in
Section~\ref{sec:momentum}. Fortunately, the use of the warm-started
power iteration to approximate the nuclear norm of $\tG_\ell$
essentially corrects this deficiency.

\begin{table}[htbp]
\centering \footnotesize
\begin{tabular}{|l|r|r|r|r|r|r|r|}
\hline
Dataset & No Mom. & {\sf MMconst} & {\sf MGconst} & {\sf MGconstA} &{\sf MMdecay} & {\sf MGdecay} & {\sf MGdecayA}\\
\hline
Fashion-MNIST  &  10 &  9 &  5 &  9 &  10 &  6 &   8 \\
MoE-FMNIST     &  42 & 34 & 18 & 34 &  28 & 21 &  27 \\
CovType        & 105 & 95 & 56 & 99 & 105 & 67 & 101 \\ 
MovieLens      &  89 & 90 & 57 & 84 &  79 & 50 &  87 \\
Criteo         &  89 & 84 & 72 & 93 &  82 & 58 &  96 \\
SVHN           &   1 &  1 &  1 &  1 &   1 &  1 &   1 \\
\hline
\end{tabular}
\caption{Throughput for asynchronous momentum variants}
\label{tab:mom-vars-tput}
\end{table}

\noindent
Finally focussing on the classification problems, we report, in
Table~\ref{tab:classif-var}, the variance of the accuracy obtained for
the 8 independent runs.
This table suggests that the variance on accuracy is nearly always better for
the {\sf MM} momentum than for the {\sf MG} approach.  In particular,
a comparison of {\sf MMconst} with {\sf MGconst} shows
that the former improves the accuracy of all classification problems
by as much as 3.58\% (for SVHN).

\begin{table}[htbp]
\centering \footnotesize
\begin{tabular}{|l|r|r|r|r|r|r|r|}
\hline
Dataset & No Mom. & {\sf MMconst} & {\sf MGconst} & {\sf MGconstA} &{\sf MMdecay} & {\sf MGdecay} & {\sf MGdecayA}\\
\hline
Fashion-MNIST  & $\pm$0.23 & $\pm$0.23 & $\pm$0.26 & $\pm$0.30 & \textbf{$\pm$0.09} & $\pm$0.18 & $\pm$0.22 \\
MoE-FMNIST     & \textbf{$\pm$0.30} & $\pm$0.33 & $\pm$0.38 & $\pm$0.34 & $\pm$0.79 & $\pm$0.38 & $\pm$0.53 \\
CovType        & \textbf{$\pm$0.11} & $\pm$0.14 & $\pm$0.28 & $\pm$0.17 & $\pm$0.18 & $\pm$0.24 & $\pm$0.20 \\
SVHN           & $\pm$2.78 & $\pm$1.15 & $\pm$1.54 & \textbf{$\pm$0.43} & $\pm$1.27 & $\pm$2.10 & $\pm$1.22 \\
\hline
\end{tabular}
\caption{Accuracy variance for asynchronous momentum variants on
  classification problems}
\label{tab:classif-var}
\end{table}

\end{document}